%% file: preprint_main.tex
\documentclass[10pt,twocolumn,letterpaper]{article}

\usepackage[pagenumbers]{cvpr}
\usepackage{fontawesome5} 
\usepackage{graphicx}    
\input{preamble}

\definecolor{cvprblue}{rgb}{0.21,0.49,0.74}
\usepackage[pagebackref,breaklinks,colorlinks,
            allcolors=cvprblue]{hyperref}

\hypersetup{
    pdftitle={MARS: Multiscale Appearance-Residual Streams
              for Video Forgery Detection},
    pdfauthor={Ahmed Y. Radwan and Shaina Raza}
}

\title{MARS: Multiscale Appearance-Residual Streams
for Video Forgery Detection}

\author{
Ahmed Y. Radwan \qquad
Shaina Raza\thanks{Corresponding author:
\href{mailto:shaina.raza@vectorinstitute.ai}
{shaina.raza@vectorinstitute.ai}.}\\
Vector Institute for Artificial Intelligence\\
Toronto, Canada\\
{\tt\small
\{ahmed.radwan,shaina.raza\}@vectorinstitute.ai}
}

\begin{document}
\maketitle

\begin{abstract}
Generalizing deepfake detection across facial manipulation and fully generated video remains challenging due to forgery traces varying with the manipulation method and scene content. Explicit spatial and temporal feature differences offer a way to capture this variation, but also respond to natural texture and motion, motivating their joint interpretation with appearance features. We investigate whether combining multiscale feature differences with adapted appearance representations improves detection across both forgery categories. We introduce Multiscale Appearance-Residual Streams (MARS), a two-path video detector. The residual path applies fixed spatial and temporal difference-of-Gaussians operators to frozen intermediate features and models the resulting differences alongside appearance context across frames. The semantic path adapts the encoder's normalization parameters for frame classification, and a clip-dependent gate combines their predictions. Under a shared protocol across seven datasets, MARS achieves the highest average AUROC of 91.43\%, outperforming seven compared detectors. When trained on facial manipulations alone, MARS achieves 93.99\% AUROC on GenVideo and 75.50\% on DF26, demonstrating transfer to generated videos without generated-video training data. MARS also provides signed patch and frame contributions, whose faithfulness we evaluate through controlled spatial and temporal removal against random controls. We provide the models, code, and a unified training and evaluation pipeline.

\smallskip
\noindent
\href{https://vectorinstitute.github.io/mars/}{\faGlobe~Project page}
\quad
\href{https://github.com/VectorInstitute/mars}{%
\raisebox{-0.2ex}{\includegraphics[height=2ex]{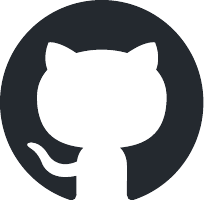}}~GitHub}
\end{abstract}

\section{Introduction}
\label{sec:introduction}

Deepfake detection (DFD) assesses video authenticity across facial manipulations, such as face swapping and reenactment~\cite{rossler2019faceforensics++}, and fully generated videos~\cite{chen2026demamba}. Their misuse enables
impersonation and misleading content~\cite{raza2026deepfakes}, requiring detectors that generalize beyond the forgeries seen during training. Existing approaches to DFD mostly examine spatial residuals~\cite{luo2021generalizing}, adjacent-frame inconsistencies~\cite{gu2021spatiotemporal}, facial style changes~\cite{choi2024exploiting}, or feature dynamics~\cite{zheng2025d3}. Pretrained visual encoders provide complementary appearance features learned through
image--text supervision~\cite{radford2021learning} or self-supervised objectives~\cite{he2022masked}.
\begin{figure}
    \centering
    \includegraphics[width=0.85\linewidth]{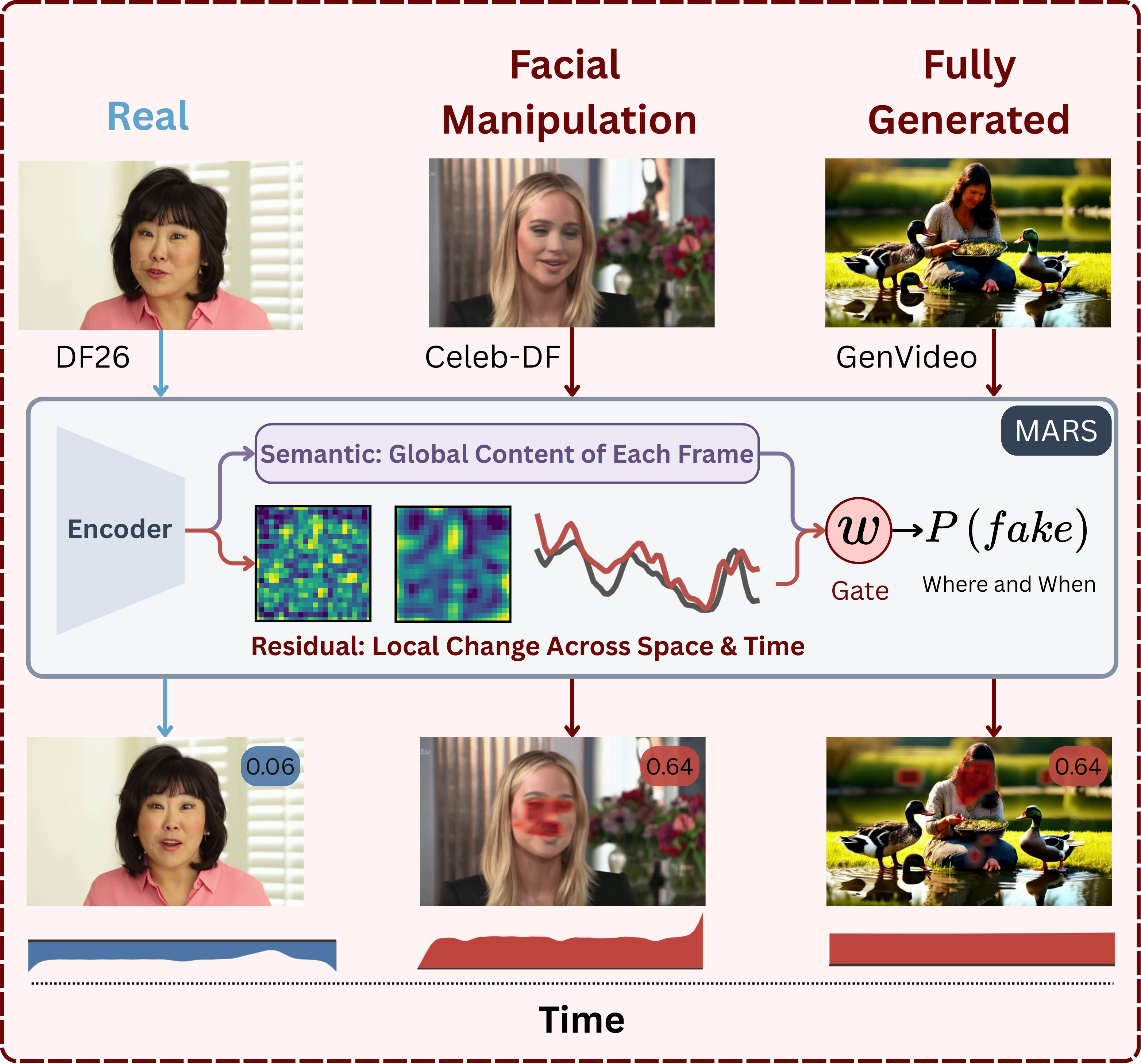}
    \caption{\textbf{One detector for facial manipulation and generated video.} A real DF26 video, a Celeb-DF facial manipulation, and a GenVideo clip pass through MARS, whose residual path compares local changes across space and time and whose semantic path reads the global content of each frame. Badges: video-level $P(\mathrm{fake})$; overlays: patch contributions to the decision; bottom: contribution over time (red: toward fake, blue: toward real).}
    \label{fig:mars-teaser}
\end{figure}

One of the common manipulation techniques is facial manipulation, which can introduce localized changes~\cite{rossler2019faceforensics++,li2020face}, whereas fully generated videos may exhibit alterations in appearance and motion across the entire frame~\cite{chen2026demamba}. This variation motivates feature comparisons across spatial neighborhoods and temporal intervals at multiple scales. Seminal works, such as UNITE studies facial manipulation and generated-video detection using pretrained features and attention diversity~\cite{kundu2025towards}. GenD demonstrates the effectiveness of adapting pretrained representations through normalization tuning~\cite{yermakov2026deepfake}. These advances raise a question: \textit{do explicit multiscale spatial and temporal feature differences add useful information to adapted appearance representation?} Residuals emphasize local variation~\cite{luo2021generalizing}, but also respond to natural texture and motion. We therefore investigate their joint use with appearance context for detection across both forgery categories.


\begin{figure*}[t]
  \centering
  \includegraphics[width=\textwidth]{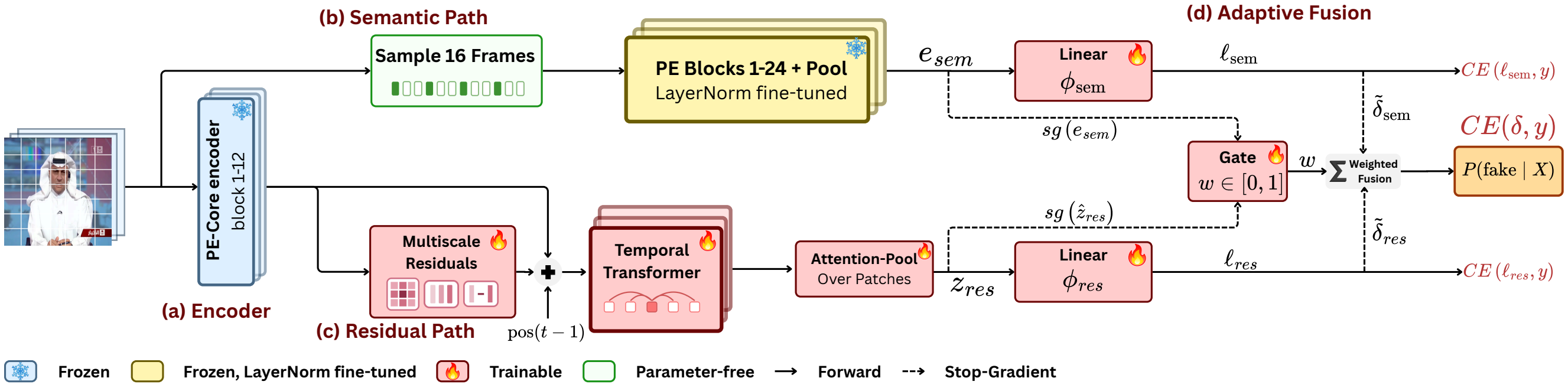}
  \caption{\textbf{Overview of MARS.} (a)~Every frame of a clip passes through the first 12 blocks of a frozen PE-Core-L encoder. (b)~The semantic path samples 16 frames and passes them through all 24 blocks and the attention pool of the same encoder, training only its own layer normalizations, then classifies each frame with a linear head $\phi_{\mathrm{sem}}$. (c)~The residual path compares the block-12 patch features across space and time at multiple scales, adds them to the appearance features with a frame-position encoding, models each patch over time with a temporal transformer, and pools over patches into $z_{\mathrm{res}}$ for a linear head $\phi_{\mathrm{res}}$. (d)~A gate predicts the weight $w$ from both clip representations and mixes the standardized margins into $P(\mathrm{fake}\mid X)$. Each path is trained by its own cross-entropy; dashed arrows stop gradients, so the fused loss trains only the gate.}
  \label{fig:mars_overview}
\end{figure*}

To investigate this complementarity, we introduce \textbf{M}ultiscale \textbf{A}ppearance-\textbf{R}esidual \textbf{S}treams (\textbf{MARS}), a full-frame detector with two complementary paths (\cref{fig:mars-teaser}). The residual path applies fixed spatial and temporal difference-of-Gaussians (DoG) operators~\cite{lowe2004distinctive} to frozen intermediate patch features, emphasizing changes at multiple scales while retaining the original features as appearance context. A temporal transformer then models how these changes develop across the clip.
The semantic path uses the complete encoder to classify selected frames, tuning only normalization parameters~\cite{basu2024strong} and a classification head while keeping the remaining encoder weights frozen. A lightweight gate combines the two paths' predictions for each clip (\cref{fig:mars_overview}). Our contributions are: (1) We introduce MARS, combining multiscale feature differences with adapted appearance representations for unified video forgery detection. (2) Our empirical analysis across seven datasets shows that, under a shared-protocol in which we retrain all compared detectors, MARS achieves 91.43\% average AUROC when trained on FF++ and GTA-V, exceeding the strongest baseline by 1.40 points with approximately $5\times$ less training compute. (3) Training on facial manipulations alone achieves 93.99\% AUROC on GenVideo~\cite{chen2026demamba} and 75.50\% on DF26~\cite{shykula2026df26}, demonstrating transfer to generated videos. (4) Ablations assess the components' complementarity, while controlled removal tests evaluate the faithfulness of signed patch and frame contributions derived from MARS. 

\section{Related Work}
\label{sec:related}

\paragraph{Facial Forgery Detection.}
Facial forgery involves altering a person's identity or expression in a video, and detectors look for forensic cues, the traces these alterations may leave in spatial appearance and across frames, with spatiotemporal inconsistency learning (STIL) combining spatial inconsistencies and adjacent-frame feature differences~\cite{gu2021spatiotemporal}, while TI$^2$Net and StyleFlow model temporal changes in facial identity and style representations, respectively~\cite{liu2023ti,choi2024exploiting}. More recent approaches adapt foundation models for generalizable detection. ForAda supervises a CLIP adapter with blending boundaries derived from manipulation masks~\cite{cui2025forensics}, whereas GenD tunes normalization layers~\cite{yermakov2026deepfake} and DFD-HR selectively processes features through layer pruning, token selection, and expert routing~\cite{sun2026dfd}. 

However, these methods use face-cropping, excluding evidence outside facial regions. This limits their direct application to generated videos that may contain no detectable faces or exhibit synthesis artifacts throughout the scene. MARS instead processes full frames and combines appearance representations with multiscale spatial and temporal feature differences.

\paragraph{Generated Video Detection and Unified Detection.}
Generated-video detectors examine synthesis traces across the full scene. DeMamba models local spatio-temporal inconsistencies in pretrained visual features~\cite{chen2026demamba}, while WaveRep uses wavelet-based training augmentation to encourage forensic learning~\cite{corvi2026seeing}. Other approaches detect deviations in feature trajectories or diffusion-derived spatiotemporal statistics~\cite{zheng2025d3,interno2026ai,zhang2026physics}, but primarily evaluate transfer across generators. UNITE addresses both facial manipulation and full-scene synthesis using FF++ and GTA-V training data, full-frame SigLIP features, and an attention-diversity loss~\cite{kundu2025towards}.
MARS shares this unified detection setting but explicitly computes spatial and temporal differences at multiple scales in frozen intermediate features. It combines these differences with adapted appearance representations to examine their complementary value across both forgery categories.


\paragraph{Low-level and Temporal Cues.}
Explicit forensic representations capture spatial and temporal variation alongside visual content. High-freq combines multiscale high-frequency features with an RGB pathway through residual-guided spatial attention~\cite{luo2021generalizing}, while D3 \cite{zheng2025d3} measures changes in inter-frame feature distances and ReStraV classifies the curvature and stepwise distances of feature trajectories~\cite{interno2026ai}.

Our focus is on jointly modeling multiscale spatial and temporal
feature differences while retaining appearance context, rather
than using spatial residuals or temporal trajectory statistics
alone.

\section{Method}
\label{sec:method}

{Given a video with a video-level real or fake label, our goal is a single full-frame detector that generalizes to facial manipulations and fully generated videos unseen during training.} MARS combines explicit spatial and temporal comparisons with pretrained appearance features. 
Its residual path models multiscale changes in intermediate patch features across the clip, while its semantic path adapts the encoder's final representation for frame-level classification. A clip-dependent gate combines their predictions, and both paths provide outputs for post-hoc inspection. We provide further details in \cref{app:impl}.

\subsection{Overview and Encoder}
\label{sec:overview}

\paragraph{Input and notation.}

Following UNITE's temporal sampling protocol~\cite{kundu2025towards}, we sample each video at stride two and divide it into
non-overlapping clips of $T=64$ frames. Each clip $X=(x_1,\ldots,x_T)$ inherits its video's label $y\in\{0,1\}$, indicating real or fake respectively, and each frame is resized to $336\times336$ pixels. Short clips are padded by repeating their last frame, with a validity mask $m\in\{0,1\}^{T}$ identifying the $T_v=\sum_t m_t$ genuine frames, which form a prefix. All temporal operations exclude padding, and video predictions average clip probabilities.

\paragraph{Frozen encoder.}
We build on perception encoder (PE-Core-L/14)~\cite{bolya2026perception}, which divides each $336\times336$ frame into a $24\times24$ grid of patches.

The resulting $N=576$ patch tokens and one class token, each of dimension $D=1024$, are processed by $L=24$ transformer blocks. We denote blocks $a$ to $b$ by $\mathcal{E}_{a:b}(\,\cdot\,;\gamma)$, where $\gamma$ contains the layer normalization (LN)~\cite{ba2016layer} scale and shift parameters. Both paths share frozen attention and multi-layer perceptron (MLP) weights but use different encoder depths and normalization parameters.

The residual path processes every frame through the first $k=12$ blocks with frozen pretrained normalization parameters $\gamma$,
\begin{equation}
Z_t=\mathcal{E}_{1:k}(x_t;\gamma)
\in\mathbb{R}^{(N+1)\times D},
\qquad t=1,\ldots,T,
\label{eq:split}
\end{equation}

and collects their patch tokens into $F\in\mathbb{R}^{T\times N\times D}$ for the residual path. The semantic path processes a subset of frames through all $L$ blocks, tuning its own normalization parameters $\gamma^{\mathrm{sem}}$, initialized from $\gamma$, throughout the encoder. 

\subsection{Multiscale Residual Path}
\label{sec:residual}

The residual path computes spatial and temporal feature residuals by subtracting patch features smoothed at consecutive
scales, following Gaussian scale-space theory~\cite{lindeberg1994scale,lindeberg2011generalized} and its DoG formulation~\cite{lowe2004distinctive}. 

\paragraph{Spatial and temporal comparisons.}
We apply LN to each patch feature and project it from $D=1024$ to $d_r=256$ channels,

\begin{equation}
\bar F=\mathrm{LN}(F),
\qquad
R=\bar F W_r+b_r\in\mathbb{R}^{T\times N\times d_r}.
\label{eq:residual_projection}
\end{equation}


We define Gaussian smoothing $G_\sigma$ over $n$ positions with row-normalized entries,

\begin{equation}
[G_\sigma]_{ij}
=
\frac{\exp\!\left(-(i-j)^2/(2\sigma^2)\right)}
{\sum_{a=1}^{n}\exp\!\left(-(i-a)^2/(2\sigma^2)\right)}.
\label{eq:gaussian}
\end{equation}


We parameterize the kernel width as $\sigma(\tau)=\sqrt{2\tau}$. The spatial operator $\mathcal{S}_{\tau}$ applies smoothing separably along the rows and columns of each frame's patch grid, while the temporal operator $\mathcal{T}_{\tau}^{m}$ applies it along the frame axis within the valid prefix, leaving padded positions unchanged. For the scales $\{\tau_j\}_{j=1}^{J}=\{0.5,1,2,4\}$, we compute consecutive-scale differences,

\begin{equation}
\begin{aligned}
r_j^{\mathrm{sp}}
&=\mathcal{S}_{\tau_j}R-\mathcal{S}_{\tau_{j+1}}R,\\
r_j^{\mathrm{tm}}
&=\mathcal{T}_{\tau_j}^{m}R-\mathcal{T}_{\tau_{j+1}}^{m}R,
\qquad j=1,\ldots,J-1.
\end{aligned}
\label{eq:bands}
\end{equation}


This produces three spatial and three temporal DoG streams. Since each row of $G_\sigma$ sums to one, subtraction cancels constant components and emphasizes variation at different spatial and temporal scales. \cref{fig:spatial_residual} shows the spatial response for one channel, \cref{fig:spatial_energy} its magnitude across channels, and \cref{fig:temporal_comparisons} the temporal responses at a fixed patch and channel.
We also compare consecutive frames using their projected patch features $R_t\in\mathbb{R}^{N\times d_r}$,

\begin{equation}
r_t^{\Delta}
=
\begin{cases}
0, & t=1,\\
m_t\,m_{t-1}\,(R_t-R_{t-1}), & t=2,\ldots,T,
\end{cases}
\label{eq:diff}
\end{equation}

where the product $m_t m_{t-1}$ keeps the difference only when both frames are genuine. The subtraction compares features at the same patch position in consecutive sampled frames and does not track object motion. Together with the six DoG streams, this gives $M=7$ residual streams, none of which contains learnable parameters.

\paragraph{Retaining appearance alongside residuals.}
The residual streams emphasize changes but suppress shared content that may help interpret them. We therefore retain the normalized patch features $\bar F$ as appearance context, project them and the normalized residual streams to a common width $d=1152$, and add the outputs. Here,
$W_a\in\mathbb{R}^{D\times d}$ and $W_j\in\mathbb{R}^{d_r\times d}$ are learned projections
(biases omitted), and each $\mathrm{LN}_j$ has its own learned scale and shift.

\begin{equation}
U
=
\underbrace{\bar F W_a}_{\text{appearance}}
+
\sum_{j=1}^{M}\underbrace{\mathrm{LN}_j(r_j)\,W_j}_{\text{residual stream } j}
\in\mathbb{R}^{T\times N\times d},
\label{eq:fusion}
\end{equation}

where $r_1,\ldots,r_M$ are the seven residual streams. Per-stream normalization reduces differences in magnitude, while the learned projections determine how each stream and the appearance features contribute to the sum.

\paragraph{Frame order.}
We add a sinusoidal frame-position encoding~\cite{vaswani2017attention}, shared across all patches of each frame,

\begin{equation}
U_t\leftarrow U_t+\mathrm{pos}(t-1),
\qquad t=1,\ldots,T_v.
\label{eq:position}
\end{equation}

Because the valid frames form a prefix, $t-1$ is also the frame's position among the valid frames. 

\paragraph{Temporal modeling and readout.}
To assess how feature changes develop across the clip, we process each patch position $n$ as a temporal sequence $(U_{1,n},\ldots,U_{T,n})$ using a temporal transformer $\mathrm{TT}$, whose attention operates across frames with padded keys masked~\cite{bertasius2021space}. We then average its outputs over valid frames to obtain one representation per patch,

\begin{equation}
u_n
=
\frac{1}{T_v}\sum_{t=1}^{T}
m_t\,\mathrm{TT}(U_{\cdot,n};m)_t
\in\mathbb{R}^{d}.
\label{eq:tpool}
\end{equation}
Finally, an attention pool~\cite{ilse2018attention} weighs the positions and forms the clip representation $z_{\mathrm{res}}$, which a linear classifier $\phi_{\mathrm{res}}$ maps to the residual logits,
\begin{equation}
\begin{aligned}
\alpha_n
&=\operatorname{softmax}_{n}
\left(a^{\top}\tanh(W_p u_n)\right),\\
z_{\mathrm{res}}&=\sum_{n=1}^{N}\alpha_n u_n,
\qquad
\ell_{\mathrm{res}}=\phi_{\mathrm{res}}(z_{\mathrm{res}})\in\mathbb{R}^{2}.
\end{aligned}
\label{eq:spool}
\end{equation}

\subsection{Semantic Path}
\label{sec:semantic}

\paragraph{Adaptation of the encoder normalization.}
The residual path uses intermediate patch features, while the semantic path uses the encoder's final representation to classify frame appearance,
\begin{equation}
q_t=\mathcal{E}_{1:L}(x_t;\gamma^{\mathrm{sem}})
\in\mathbb{R}^{D},
\label{eq:semantic_encoder}
\end{equation}
where $\mathcal{E}_{1:L}$ includes the final normalization and attention-pooling head. Following GenD~\cite{yermakov2026deepfake}, we tune only the LN scale and shift parameters $\gamma^{\mathrm{sem}}$ throughout the encoder~\cite{farina2025rethinking}, including its input and readout normalizations, while keeping attention and MLP weights frozen. Separate LN parameters in blocks $1$ to $k$ preserve the residual features $Z_t$, requiring an additional pass through these blocks for the selected semantic frames.

\paragraph{Frame representations and clip prediction.}

To limit the cost of its full encoder pass, the semantic path processes $T_{\mathrm{sem}}=16$ frames per clip, with indices spread evenly over the valid frames~\cite{wang2016temporal},
\begin{equation}
t_i
=
1+\operatorname{round}
\left(\frac{i-1}{T_{\mathrm{sem}}-1}\,(T_v-1)\right),
\qquad i=1,\ldots,T_{\mathrm{sem}}.
\label{eq:semantic_sampling}
\end{equation}
When $T_v<T_{\mathrm{sem}}$, some frames are selected more than once. We normalize each feature to unit length and apply a linear classifier $\phi_{\mathrm{sem}}$, then average the frame logits and normalized features to obtain $\ell_{\mathrm{sem}}$ and $e_{\mathrm{sem}}$,
respectively,
\begin{equation}
\begin{aligned}
\hat q_i &= \frac{q_{t_i}}{\|q_{t_i}\|_2},
\qquad
e_{\mathrm{sem}} = \frac{1}{T_{\mathrm{sem}}}\sum_{i=1}^{T_{\mathrm{sem}}}\hat q_i,\\
\ell_{\mathrm{sem}} &= \frac{1}{T_{\mathrm{sem}}}\sum_{i=1}^{T_{\mathrm{sem}}}\phi_{\mathrm{sem}}(\hat q_i)\in\mathbb{R}^{2}.
\end{aligned}
\label{eq:semantic}
\end{equation}

\subsection{Adaptive Fusion and Training}
\label{sec:fusion}
\begin{table*}[t]
\centering\small
\setlength{\tabcolsep}{3.2pt}
\renewcommand{\arraystretch}{0.85}
\caption{\textbf{Cross-dataset detection under two training settings.}
Video-level AUROC (\%) and EER (\%) for methods retrained under
the shared protocol with a fixed seed and three epochs. Checkmarks indicate training corpora. Avg.\ is the equal mean over the seven evaluation datasets. Best results are \textbf{bold} and second-best results are \underline{underlined} within each training setting, including ties. AP is reported in \cref{tab:ap-main}.
}
\label{tab:crossdataset_main_table}
\resizebox{\textwidth}{!}{%
\begin{tabular}{@{}ccl*{16}{c}@{}}
\toprule
\multicolumn{2}{c}{\textbf{Train}} & &
\multicolumn{2}{c}{\textbf{Source test}} &
\multicolumn{8}{c}{\textbf{Facial manipulation}} &
\multicolumn{4}{c}{\textbf{Generated video}} &
\multicolumn{2}{c}{\textbf{Avg.}}\\
\cmidrule(lr){1-2}
\cmidrule(lr){4-5}
\cmidrule(lr){6-13}
\cmidrule(lr){14-17}
\cmidrule(lr){18-19}
FF++ & GTA-V & &
\multicolumn{2}{c}{FF++} &
\multicolumn{2}{c}{CelebDF} &
\multicolumn{2}{c}{DF-TIMIT} &
\multicolumn{2}{c}{UADFV} &
\multicolumn{2}{c}{HiFiFace} &
\multicolumn{2}{c}{GenVideo} &
\multicolumn{2}{c}{DF26} & & \\
\cmidrule(lr){4-5}
\cmidrule(lr){6-7}
\cmidrule(lr){8-9}
\cmidrule(lr){10-11}
\cmidrule(lr){12-13}
\cmidrule(lr){14-15}
\cmidrule(lr){16-17}
& & Method &
AUROC$\uparrow$ & EER$\downarrow$ &
AUROC$\uparrow$ & EER$\downarrow$ &
AUROC$\uparrow$ & EER$\downarrow$ &
AUROC$\uparrow$ & EER$\downarrow$ &
AUROC$\uparrow$ & EER$\downarrow$ &
AUROC$\uparrow$ & EER$\downarrow$ &
AUROC$\uparrow$ & EER$\downarrow$ &
AUROC$\uparrow$ & EER$\downarrow$\\
\midrule
& & \multicolumn{17}{l}{\textit{Frame-level}}\\
\cmark & \xmark & ForAda~\cite{cui2025forensics}
& 86.67 & 21.61
& 86.66 & 22.28 & 96.67 & 9.53
& 95.29 & \underline{8.16} & 71.26 & 35.00
& 54.54 & 46.59 & 39.53 & 57.60
& 75.80 & 28.68\\
\cmark & \xmark & Effort~\cite{yan2024effort}
& 72.28 & 34.46
& 59.50 & 40.95 & 82.19 & 24.69
& 94.00 & 17.35 & 58.28 & 44.29
& 51.10 & 49.28 & 36.58 & 57.63
& 64.85 & 38.38\\
\cmark & \xmark & GenD~\cite{yermakov2026deepfake}
& 70.42 & 34.55
& 66.13 & 39.50 & 98.91 & 5.08
& 51.10 & 53.06 & 38.76 & 56.07
& 64.32 & 40.55 & 59.04 & 43.51
& 64.10 & 38.90\\
\cmark & \xmark & DFD-HR~\cite{sun2026dfd}
& \textbf{99.80} & \textbf{1.52}
& \textbf{94.88} & \textbf{13.36}
& \textbf{99.95} & \textbf{0.94}
& 97.21 & 10.20
& \textbf{86.75} & \textbf{19.64}
& 66.16 & 38.94
& \underline{70.87} & \underline{35.46}
& \underline{87.95} & \underline{17.15}\\
\cmidrule(lr){3-19}
& & \multicolumn{17}{l}{\textit{Clip-level}}\\
\cmark & \xmark & UNITE~\cite{kundu2025towards}
& 90.22 & 16.61
& 81.15 & 24.85 & 91.16 & 16.09
& \underline{97.38} & \textbf{7.14}
& 72.07 & 33.57
& \underline{66.36} & \underline{38.81}
& 41.29 & 55.75 & 77.09 & 27.55\\
\cmark & \xmark & DeMamba~\cite{chen2026demamba}
& 51.54 & 48.30
& 47.74 & 51.56 & 49.33 & 50.39
& 49.25 & 52.04 & 50.51 & 48.93
& 51.29 & 48.73 & 48.78 & 50.64
& 49.78 & 50.09\\
\cmark & \xmark & ReStraV~\cite{interno2026ai}
& 52.80 & 50.18
& 49.00 & 51.99 & 4.25 & 88.67
& 52.04 & 48.98 & 53.44 & 47.50
& 6.92 & 85.74 & 59.82 & 43.79
& 39.75 & 59.55\\
\cmidrule(lr){3-19}
\rowcolor{oursrow}
\cmark & \xmark & \textbf{MARS (Ours)}
& \underline{99.23} & \underline{4.20}
& \underline{91.25} & \underline{16.23}
& \underline{99.85} & \underline{1.88}
& \textbf{97.92} & \underline{8.16}
& \underline{84.92} & \underline{22.14}
& \textbf{93.99} & \textbf{11.52}
& \textbf{75.50} & \textbf{29.06}
& \textbf{91.81} & \textbf{13.31}\\
\midrule
& & \multicolumn{17}{l}{\textit{Frame-level}}\\
\cmark & \cmark & ForAda~\cite{cui2025forensics}
& 86.55 & 22.32
& 85.25 & 21.41 & 96.57 & 10.23
& 95.59 & 10.20 & 73.45 & 33.57
& 64.70 & 38.82 & 40.04 & 55.32
& 77.45 & 27.41\\
\cmark & \cmark & Effort~\cite{yan2024effort}
& 71.77 & 35.09
& 59.05 & 41.82 & 82.33 & 25.62
& 92.71 & 15.31 & 57.93 & 43.57
& 56.45 & 45.41 & 37.98 & 57.93
& 65.46 & 37.82\\
\cmark & \cmark & GenD~\cite{yermakov2026deepfake}
& 70.47 & 34.38
& 67.44 & 37.51 & 98.69 & 5.39
& 51.81 & 47.96 & 47.47 & 51.43
& 64.92 & 39.64 & 59.63 & 44.21
& 65.78 & 37.22\\
\cmark & \cmark & DFD-HR~\cite{sun2026dfd}
& \textbf{99.80} & \textbf{1.43}
& \textbf{93.89} & \textbf{13.93}
& \underline{99.59} & \underline{1.80}
& 96.25 & 10.20
& \textbf{86.26} & \textbf{20.71}
& \underline{79.48} & \underline{27.69}
& \underline{74.95} & \underline{31.39}
& \underline{90.03} & \underline{15.31}\\
\cmidrule(lr){3-19}
& & \multicolumn{17}{l}{\textit{Clip-level}}\\
\cmark & \cmark & UNITE~\cite{kundu2025towards}
& 84.07 & 24.11
& 73.48 & 31.04 & 86.54 & 22.27
& \textbf{98.58} & \textbf{5.10}
& 66.74 & 38.21 & 66.78 & 34.84
& 42.64 & 55.73 & 74.12 & 30.18\\
\cmark & \cmark & DeMamba~\cite{chen2026demamba}
& 51.40 & 49.11
& 47.85 & 52.30 & 49.30 & 51.02
& 51.25 & 50.00 & 50.42 & 50.00
& 49.93 & 50.11 & 52.67 & 46.20
& 50.40 & 49.82\\
\cmark & \cmark & ReStraV~\cite{interno2026ai}
& 52.79 & 48.66
& 50.90 & 50.28 & 4.16 & 88.75
& 50.94 & 53.06 & 53.11 & 48.93
& 9.79 & 82.82 & 59.93 & 43.36
& 40.23 & 59.41\\
\cmidrule(lr){3-19}
\rowcolor{oursrow}
\cmark & \cmark & \textbf{MARS (Ours)}
& \underline{98.98} & \underline{4.46}
& \underline{91.22} & \underline{14.80}
& \textbf{99.95} & \textbf{0.94}
& \underline{96.58} & \underline{8.16}
& \underline{84.05} & \underline{22.50}
& \textbf{93.44} & \textbf{12.20}
& \textbf{75.81} & \textbf{29.78}
& \textbf{91.43} & \textbf{13.26}\\
\bottomrule
\end{tabular}%
}
\end{table*}
\paragraph{Normalization of path scores.}

The two paths can produce logits of different magnitudes, so direct fusion could favor one path because of its scale. We therefore standardize each fake-versus-real margin
$\delta_\star=\ell_\star^{(1)}-\ell_\star^{(0)}$,
for $\star\in\{\mathrm{sem},\mathrm{res}\}$,
\begin{equation}
\tilde\delta_{\star}
=
\frac{\delta_{\star}-\mu_{\star}}{\sqrt{v_{\star}+\epsilon}}.
\label{eq:standardize}
\end{equation}
Here, $\mu_{\star}$ and $v_{\star}$ are the batch mean and variance during training and their running estimates during inference, equivalent to batch normalization~\cite{ioffe2015batch} without learnable affine parameters. This makes the margins comparable in scale without calibrating their probabilities.

\paragraph{Input-dependent fusion.}
To allow the contribution of each path to vary across clips, we predict a mixture weight from their clip representations, $e_{\mathrm{sem}}$ and the unit-length residual representation $\hat z_{\mathrm{res}}=z_{\mathrm{res}}/\|z_{\mathrm{res}}\|_2$,

\begin{equation}
\begin{aligned}
h &= \big[\mathrm{sg}(e_{\mathrm{sem}});\, \mathrm{sg}(\hat z_{\mathrm{res}})\big], \\
w &= \sigma(g^{\top}h + b_g), \\
\delta &= w\,\mathrm{sg}(\tilde\delta_{\mathrm{sem}}) + (1-w)\,\mathrm{sg}(\tilde\delta_{\mathrm{res}}),
\end{aligned}
\label{eq:gate}
\end{equation}
where $\mathrm{sg}(\cdot)$ stops gradients. We initialize $g$ and $b_g$ to zero, giving both paths equal weight. Detaching the representations and margins ensures that the fusion objective trains only the gate. The clip's fake probability is


\begin{equation}
P(\mathrm{fake}\mid X)=\operatorname{sigmoid}(\delta).
\label{eq:probability}
\end{equation}

\paragraph{Training objective and video prediction.}
Each path is supervised separately, while the gate is trained through the fused prediction:
\begin{equation}
\mathcal{L}
=\mathrm{CE}(\ell_{\mathrm{sem}},y)
+\mathrm{CE}(\ell_{\mathrm{res}},y)
+\mathrm{CE}\!\left([-\delta/2,\;\delta/2],y\right),
\label{eq:loss}
\end{equation}
where $\mathrm{CE}$ denotes cross-entropy. The first two terms update their respective paths. Because the representations and margins are detached in Eq.~\eqref{eq:gate}, the final term updates only the gate. The logits $[-\delta/2,\delta/2]$ give a fake-class probability of $\operatorname{sigmoid}(\delta)$.

\subsection{Post-hoc Spatial and Temporal Evidence}
\label{sec:evidence}
We inspect which regions and frames support MARS's prediction without an additional explanation model. We decompose the fused margin into signed frame scores, combining the residual classifier's patch contributions with the semantic frame margins, so that their sum, with bias and normalization offsets, recovers the fused margin. These scores show which features the classifier rely on, not which pixels caused the prediction. Summing the residual contributions over time also gives a $24\times24$ patch map per clip. We evaluate the faithfulness of this evidence through deletion tests~\cite{petsiuk2018rise,rong22consistent}, measuring whether removing highly ranked patches or frames changes predictions more than equal-budget random removal. \cref{app:evidence} provides the score definitions and evaluation details.
\begin{table*}[t]
\centering\scriptsize
\setlength{\tabcolsep}{2.5pt}
\renewcommand{\arraystretch}{0.8}
\caption{\textbf{Per-generator detection on GenVideo and DF26.}
Video-level AUROC (\%) for methods trained on FF++ + GTA-V
under the shared protocol with a fixed seed and three epochs.
Each generator is evaluated against all authentic videos
from the same dataset.
Best results are \textbf{bold} and second-best results are
\underline{underlined} in each column.
Per-generator EER is reported in \cref{tab:ap-pergen-detail-eer}.}
\label{tab:recent_per_generator}
\resizebox{\textwidth}{!}{%
\begin{tabular}{@{}l*{20}{c}@{}}
\toprule
& \multicolumn{10}{c}{\textbf{GenVideo}}
& \multicolumn{10}{c}{\textbf{DF26}}\\
\cmidrule(lr){2-11}\cmidrule(lr){12-21}
& \multicolumn{10}{c}{}
& \multicolumn{4}{c}{\textit{Commercial}}
& \multicolumn{6}{c}{\textit{Open-source}}\\
\cmidrule(lr){12-15}\cmidrule(lr){16-21}
Method & Sora & \shortstack{Morph\\Studio}
& \shortstack{Runway\\Gen2} & HotShot & Lavie & Show-1
& \shortstack{Moon\\Valley} & Crafter
& \shortstack{Model\\Scope} & \shortstack{Wild\\Scrape}
& \shortstack{Grok\\Imagine}
& \shortstack{Kling\\3.0}
& \shortstack{Veo\\3.1}
& \shortstack{Wan\\2.6}
& \shortstack{HV1.5\\I2V}
& \shortstack{HV1.5\\T2V}
& \shortstack{Wan2.2\\I2V}
& \shortstack{Wan2.2\\T2V}
& \shortstack{LTX2.3\\I2V}
& \shortstack{LTX2.3\\T2V}\\
\midrule
\grp{21}{Frame-level}
ForAda
& 59.05 & 73.09 & 46.58 & \underline{72.77}
& 65.08 & 65.49 & 57.70 & 66.61
& 79.72 & 70.96
& 19.68 & 30.60 & 14.16 & 30.90
& 40.99 & 58.11 & 60.62 & 42.36
& 53.38 & 32.36\\
Effort
& 46.47 & 63.11 & 47.72 & 58.61
& 55.66 & 61.04 & 47.06 & 52.52
& 69.30 & 67.16
& 26.74 & 44.56 & 20.91 & 30.30
& 45.87 & 42.59 & 49.79 & 36.69
& 47.47 & 27.08\\
GenD
& 66.36 & 61.41 & 76.76 & 43.09
& 58.02 & 47.56 & 73.53 & 78.56
& 63.36 & 64.50
& \textbf{78.63} & 74.64 & 52.40 & 58.52
& 51.14 & 63.36 & 53.34 & 50.10
& 60.23 & 60.85\\
DFD-HR
& \underline{71.22} & \underline{81.57}
& \underline{85.03} & 57.81
& \underline{73.93} & \underline{76.17}
& 66.71 & \underline{93.46}
& \underline{86.70} & \underline{79.41}
& 59.65 & \underline{81.42}
& \underline{63.98} & \underline{77.26}
& \underline{62.85} & \textbf{87.61}
& \textbf{80.27} & 65.79
& \textbf{89.92} & \textbf{76.16}\\
\midrule
\grp{21}{Clip-level}
UNITE
& 55.80 & 70.01 & 51.94 & 67.86
& 70.20 & 73.87 & \underline{77.32} & 66.09
& 66.41 & 71.47
& 44.75 & 46.26 & 29.80 & 39.97
& 43.66 & 57.67 & 50.79 & 34.07
& 44.01 & 32.86\\
DeMamba
& 55.49 & 43.22 & 42.93 & 61.39
& 51.06 & 57.97 & 42.61 & 47.33
& 55.37 & 55.12
& 54.80 & 57.40 & 50.93 & 47.39
& 50.60 & 26.19 & 50.95 & 61.56
& 52.30 & \underline{74.54}\\
ReStraV
& 1.60 & 23.08 & 0.32 & 4.74
& 13.11 & 11.74 & 0.50 & 12.12
& 13.57 & 12.52
& 66.24 & 65.00 & 48.29 & 63.33
& 48.66 & 48.56 & 67.57 & \underline{70.00}
& 54.28 & 68.28\\
\midrule
\rowcolor{oursrow}
\textbf{MARS (Ours)}
& \textbf{80.56} & \textbf{93.01}
& \textbf{94.40} & \textbf{85.52}
& \textbf{93.32} & \textbf{87.31}
& \textbf{96.07} & \textbf{98.30}
& \textbf{95.50} & \textbf{93.18}
& \underline{72.20} & \textbf{81.57}
& \textbf{78.12} & \textbf{84.45}
& \textbf{74.45} & \underline{86.49}
& \underline{72.19} & \textbf{72.95}
& \underline{73.37} & 65.89\\
\bottomrule
\end{tabular}}
\end{table*}
\section{Experimental Setup}
\label{sec:setup}

\subsection{Data, Training, and Selection}
We train on FF++ c23~\cite{rossler2019faceforensics++} and GTA-V~\cite{hu2021sail}, and evaluate out-of-domain generalization across facial manipulation and full-scene synthesis using CelebDF-V2~\cite{li2020celeb}, DeepFakeTIMIT~\cite{korshunov2018deepfakesdftmit}, HiFiFace~\cite{ijcai2021-157-hififace}, UADFV~\cite{yang2019exposing}, GenVideo~\cite{chen2026demamba}, and DF26~\cite{shykula2026df26}.
After each epoch, we evaluate on the FF++ validation split and select the checkpoint with the highest video-level AUROC, then evaluate it on the source test splits and out-of-domain datasets. \cref{app:datasets} summarizes the datasets, with their splits and handling.
All methods are trained with a fixed seed for three epochs using AdamW, a learning rate of $10^{-4}$, weight decay of $10^{-4}$, and an effective batch size of 32 clips. The learning rate is halved every 1,000 optimizer updates. The complete shared training settings are provided in \cref{app:protocol}.

\subsection{Baselines and Evaluation Protocol}
We compare MARS with seven methods covering different approaches
to forgery detection.
ForAda~\cite{cui2025forensics},
Effort~\cite{yan2024effort},
GenD~\cite{yermakov2026deepfake}, and
DFD-HR~\cite{sun2026dfd} adapt pretrained image representations,
while DeMamba~\cite{chen2026demamba} and
ReStraV~\cite{interno2026ai} model video features for
generated-video detection.
UNITE~\cite{kundu2025towards} provides a unified video detector
for facial manipulation and full-scene synthesis.
\cref{tab:methods} summarizes their backbones, adaptation
strategies, objectives, and trainable parameter counts.
All methods use the input pipeline described in \cref{sec:overview}. 

Frames are resized to each method's input resolution, scaled to $[-1,1]$, and converted by it's encoder wrapper. For every detector, a video's fake score is the mean of its clip probabilities. Implementation details for each method are provided in \cref{app:methods}. We examine alternative training settings for GenD and DeMamba in \cref{tab:ap-fairness}.

We report video-level AUROC, EER, and AP (\%), treating fake videos as the positive class. For GenVideo and DF26, each generator is evaluated against the same authentic-video pool within its dataset. Macro results weight generators equally, while pooled results count each video once.


\subsection{Experimental Results}
\label{sec:results}

\subsubsection{Overall Evaluation Results}

\textbf{Generalization to Unseen Datasets.}
Training on facial manipulations alone allows MARS to transfer to full-scene synthesis, achieving 93.99\% AUROC on GenVideo and 75.50\% on DF26, compared with 66.16\% and 70.87\% for DFD-HR (\cref{tab:crossdataset_main_table}). Although DFD-HR remains stronger on several facial datasets, MARS achieves the best overall results with 91.81\% average AUROC and 13.31\% EER. Adding GTA-V improves generated-video detection for several baselines, particularly DFD-HR, whose GenVideo AUROC increases to 79.48\%. MARS remains close to its FF++ only results, reaching 91.43\% average AUROC and 13.26\% EER, suggesting that its transfer across both forgery categories does not depend on GTA-V training.
\textbf{Per-Generator Results.}
Under FF++ + GTA-V training, MARS achieves the highest
GenVideo AUROC in all ten generator groups, averaging
91.72\% across generators compared with 77.20\% for DFD-HR
(\cref{tab:recent_per_generator}). This advantage is less consistent on DF26, where MARS leads in five groups and averages 76.17\% against DFD-HR's 74.49\%. MARS performs better on Veo~3.1 and Wan~2.6, while DFD-HR leads on both LTX2.3 modes, showing that the overall gain does not extend uniformly to every generation method. 

Overall, MARS generalizes across both forgery categories under either training setting, although its advantage varies across generators. Training and testing compute, energy use, and carbon emissions are reported in \cref{tab:footprint}.

\subsubsection{Ablation Study}
We evaluate the contribution of each path, residual stream family, semantic normalization tuning, and different encoder depths and fusion under matched training settings (\cref{tab:ablation-components}).

\paragraph{Complementarity of the Two Paths.}
Although the semantic path performs better on its own, adding the residual path raises facial AUROC by ~3.0\% and overall AUROC from 90.24\% to 91.43\%. Generated-video AUROC falls from 87.14\% to 84.63\%, showing that the paths provide complementary information for facial detection. The overall gain also holds when averaged over five seeds (\cref{tab:ap-fairness}). 

\begin{table}[t]
\centering\footnotesize
\setlength{\tabcolsep}{3pt}
\caption{\textbf{Component ablation of MARS.}
AUROC (\%) under the shared training protocol.
Sem./Res.: semantic/residual paths; App./Spat./Temp./FD:
appearance, spatial DoG, temporal DoG, and first difference;
LN: semantic LayerNorm tuning.
Facial (F) and Generated (G) average the four external facial
and two generated-video datasets, respectively.
Avg.\ includes all seven datasets.
Best in \textbf{bold}, second best \underline{underlined}.
Per-dataset results are given in \cref{tab:ap-ablation-training},
and the split layer and backbone in \cref{tab:ap-split-backbone}.}
\label{tab:ablation-components}
\resizebox{\linewidth}{!}{%
\begin{tabular}{@{}lcc cccc c ccc@{}}
\toprule
& \multicolumn{2}{c}{Path}
& \multicolumn{4}{c}{Residual-path features}
& Norm
& \multicolumn{3}{c}{AUROC$\uparrow$}\\
\cmidrule(lr){2-3}
\cmidrule(lr){4-7}
\cmidrule(lr){8-8}
\cmidrule(lr){9-11}
Variant & Sem. & Res. & App. & Spat. & Temp. & FD
& LN & F & G & Avg.\\
\midrule
\rowcolor{oursrow}
\textbf{MARS (Ours)}
& \cmark & \cmark
& \cmark & \cmark & \cmark & \cmark
& \cmark
& \underline{92.95} & \underline{84.63} & \textbf{91.43}\\
\midrule
Residual path only
& \xmark & \cmark
& \cmark & \cmark & \cmark & \cmark
& --
& 89.13 & 69.50 & 84.80\\
Semantic path only
& \cmark & \xmark
& -- & -- & -- & --
& \cmark
& 89.89 & \textbf{87.14} & 90.24\\
\midrule
w/o appearance
& \cmark & \cmark
& \xmark & \cmark & \cmark & \cmark
& \cmark
& \textbf{93.69} & 82.56 & \underline{91.28}\\
w/o spatial DoG
& \cmark & \cmark
& \cmark & \xmark & \cmark & \cmark
& \cmark
& 85.54 & 73.00 & 83.96\\
w/o temporal DoG
& \cmark & \cmark
& \cmark & \cmark & \xmark & \cmark
& \cmark
& 88.77 & 74.39 & 86.07\\
w/o first difference
& \cmark & \cmark
& \cmark & \cmark & \cmark & \xmark
& \cmark
& 92.86 & 81.57 & 90.52\\
\midrule
Frozen semantic LNs
& \cmark & \cmark
& \cmark & \cmark & \cmark & \cmark
& \xmark
& 89.87 & 52.37 & 80.39\\
\bottomrule
\end{tabular}%
}
\end{table}

\paragraph{Contribution of Residual Streams.}
Removing spatial DoG causes the largest decrease in average AUROC, from 91.43\% to 83.96\%, while removing temporal DoG lowers it to 86.07\%. Both comparisons therefore contribute beyond the retained appearance features and first difference. Removing the first difference has a smaller effect, yielding 90.52\%, whereas removing appearance produces 91.28\%, with improved facial detection but weaker generated-video detection. These results suggest that appearance provides useful context for synthesis detection, while spatial and temporal DoG account for the larger gains. We further compare fewer scales and no residual streams in \cref{tab:ap-streams-scales}.
\\\textbf{Impact of Semantic LayerNorm Tuning.}
Freezing the semantic normalization parameters reduces generated-video AUROC from 84.63\% to 52.37\%, compared with a smaller decrease from 92.95\% to 89.87\% on facial datasets. This indicates that adapting the pretrained representation through normalization is particularly important for transfer to generated videos. 
\\\textbf{Impact of Split Layer and Path Fusion.}
Comparing residual extraction depths shows that block 12 performs best overall, while deeper features favor facial detection (\cref{tab:ab-split-fusion}). Combining both paths improves overall detection, supporting their complementary value, while the learned gate matches fixed averaging (\cref{tab:ab-weight-fusion}); comparisons with additional fixed weights are reported in \cref{tab:ap-gate-fixed}.

\begin{table}[t]
\centering\footnotesize
\setlength{\tabcolsep}{3pt}
\caption{\textbf{Split layer and path fusion.} AUROC (\%), averaged over the four facial (F) and two generated-video (G) test sets, and over all seven (Avg.). In (b), fixed weights replace the gate at inference in the same trained model. Default shaded; best results are \textbf{bold} in each column. Full results in \cref{tab:ap-split-backbone,tab:ap-gate-fixed}.}
\label{tab:ab-split-fusion}

\begin{subtable}[b]{0.4\linewidth}\centering
\begin{tabular}{@{}lccc@{}}
\toprule
Block & F & G & Avg.\\
\midrule
6  & 90.0 & 81.0 & 88.6\\
\rowcolor{oursrow} 12 & 93.0 & \textbf{84.6} & \textbf{91.4}\\
18 & \textbf{94.1} & 77.0 & 89.9\\
24 & 93.6 & 80.8 & 90.7\\
\bottomrule
\end{tabular}
\caption{Split layer.}\label{tab:split}
\end{subtable}\hfill
\begin{subtable}[b]{0.58\linewidth}\centering
\begin{tabular}{@{}lccc@{}}
\toprule
Fusion weight & F & G & Avg.\\
\midrule
$w{=}0$ (residual only) & 90.6 & 76.2 & 87.7\\
$w{=}0.5$ & \textbf{93.0} & 84.3 & \textbf{91.4}\\
$w{=}1$ (semantic only) & 89.4 & 83.8 & 89.0\\
\rowcolor{oursrow} Learned $w$ & \textbf{93.0} & \textbf{84.6} & \textbf{91.4}\\
\bottomrule
\end{tabular}
\caption{Path fusion.}\label{tab:ab-weight-fusion}
\end{subtable}
\end{table}

\begin{table}[t]
\centering
\caption{\textbf{Stability across training seeds.} Video-level AUROC (\%),
mean $\pm$ std over 5 seeds, trained on FF++ + GTA-V.
Mean is the equal average over the seven test sets.
Best results are \textbf{bold} in each column.}
\label{tab:ablation-training-seeds}
\setlength{\tabcolsep}{2pt}
\resizebox{\columnwidth}{!}{%
\begin{tabular}{l cccccccc}
\toprule
Method & FF++ & CelebDF & DF-TIMIT & UADFV & HiFiFace & GenVideo & DF26 & Mean \\
\midrule
\multicolumn{9}{l}{\textit{Frame-level}} \\
ForAda  & \pmstd{86.40}{0.62} & \pmstd{84.05}{1.65} & \pmstd{95.37}{1.30} & \pmstd{95.74}{0.56} & \pmstd{72.19}{0.87} & \pmstd{68.30}{2.69} & \pmstd{40.37}{0.72} & \pmstd{77.49}{0.31} \\
Effort  & \pmstd{71.39}{0.46} & \pmstd{60.54}{2.89} & \pmstd{87.52}{3.21} & \pmstd{91.48}{0.87} & \pmstd{57.49}{1.05} & \pmstd{51.74}{9.04} & \pmstd{38.60}{3.54} & \pmstd{65.54}{1.98} \\
GenD    & \pmstd{70.37}{7.34} & \pmstd{68.17}{4.63} & \pmstd{98.44}{1.32} & \pmstd{53.15}{15.51} & \pmstd{48.91}{11.60} & \pmstd{63.26}{3.53} & \pmstd{61.51}{3.09} & \pmstd{66.26}{5.62} \\
DFD-HR  & \pmstd{\textbf{99.75}}{0.06} & \pmstd{\textbf{93.45}}{0.65} & \pmstd{99.28}{0.37} & \pmstd{97.30}{1.49} & \pmstd{85.07}{2.66} & \pmstd{80.41}{4.25} & \pmstd{71.78}{3.26} & \pmstd{89.58}{0.39} \\
\midrule
\multicolumn{9}{l}{\textit{Clip-level}} \\
UNITE   & \pmstd{90.68}{4.50} & \pmstd{79.14}{3.45} & \pmstd{91.99}{3.56} & \pmstd{\textbf{97.83}}{1.08} & \pmstd{75.37}{5.88} & \pmstd{66.96}{4.27} & \pmstd{39.05}{2.91} & \pmstd{77.29}{1.97} \\
DeMamba & \pmstd{51.74}{0.30} & \pmstd{48.40}{3.75} & \pmstd{49.59}{1.36} & \pmstd{51.84}{1.10} & \pmstd{50.94}{0.48} & \pmstd{52.36}{3.60} & \pmstd{48.15}{2.98} & \pmstd{50.43}{0.66} \\
ReStraV & \pmstd{52.99}{0.14} & \pmstd{51.00}{0.56} & \pmstd{8.87}{5.33} & \pmstd{50.23}{0.58} & \pmstd{53.36}{0.41} & \pmstd{7.42}{2.27} & \pmstd{58.60}{0.93} & \pmstd{40.35}{0.59} \\
\midrule
MARS (Ours) & \pmstd{98.94}{0.15} & \pmstd{90.56}{0.55} & \pmstd{\textbf{99.85}}{0.12} & \pmstd{97.75}{0.87} & \pmstd{\textbf{85.43}}{1.21} & \pmstd{\textbf{88.91}}{4.54} & \pmstd{\textbf{73.37}}{2.71} & \pmstd{\textbf{90.69}}{1.02} \\
\bottomrule
\end{tabular}}
\end{table}

\begin{figure}[t]
  \centering
  \includegraphics[width=0.9\columnwidth]{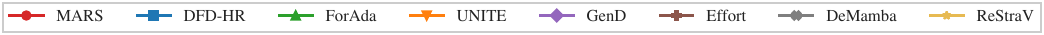}\\[2pt]
  \includegraphics[width=0.9\columnwidth]{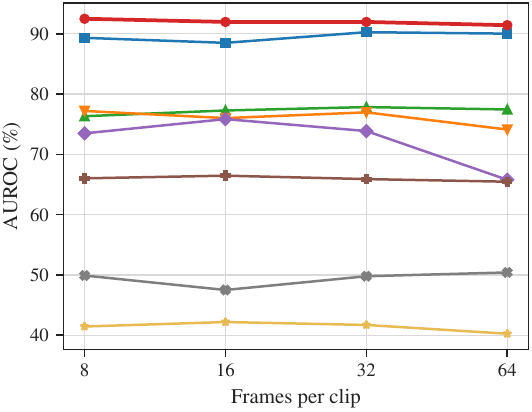}
  \caption{\textbf{Stability across clip lengths.}
    Video AUROC averaged over the seven test sets when training and testing with
    8, 16, 32 or 64 frames per clip; 64 is the main setting.
    Training length for all methods is given in \cref{tab:training-fg}, per-dataset
    results for MARS across training length and clip length in \cref{tab:duration},
    and the number of frames sampled by the semantic path in \cref{tab:semantic-frames}.}
  \label{fig:ablation-training-frames}
\end{figure}

\begin{figure*}[t]
  \centering
 \includegraphics[width=0.75\textwidth]{figures/legend_methods.pdf}\\[2pt]
  \begin{subfigure}[t]{0.249\textwidth}
    \centering
    \includegraphics[width=\linewidth]{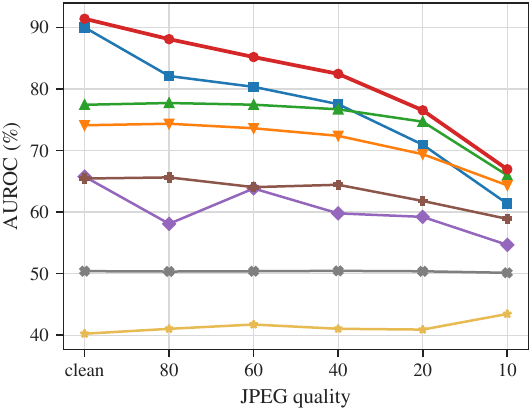}
    \caption{JPEG compression}
    \label{fig:robustness-jpeg}
  \end{subfigure}\hfill
  \begin{subfigure}[t]{0.2459\textwidth}
    \centering
    \includegraphics[width=\linewidth]{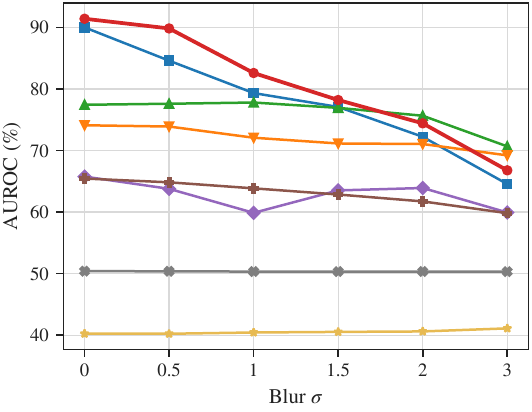}
    \caption{Gaussian blur}
    \label{fig:robustness-blur}
  \end{subfigure}\hfill
  \begin{subfigure}[t]{0.2459\textwidth}
    \centering
    \includegraphics[width=\linewidth]{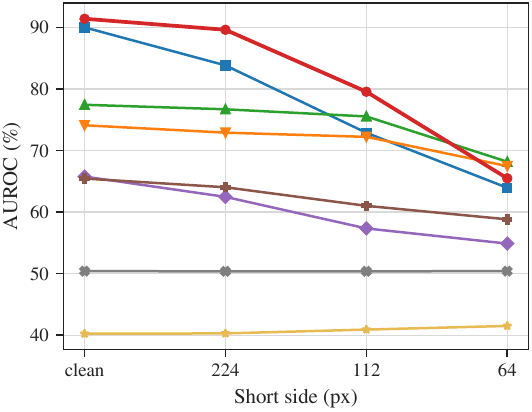}
    \caption{Downscaling}
    \label{fig:robustness-resize}
  \end{subfigure}\hfill
  \begin{subfigure}[t]{0.2459\textwidth}
    \centering
    \includegraphics[width=\linewidth]{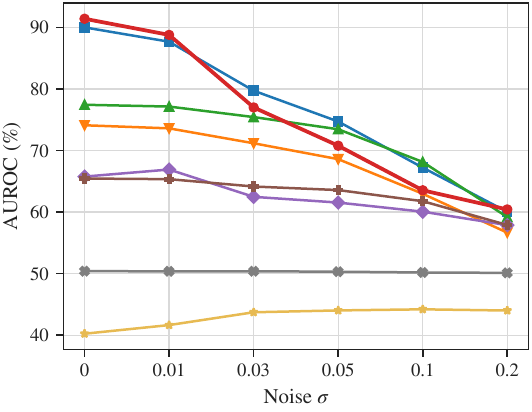}
    \caption{Gaussian noise}
    \label{fig:robustness-noise}
  \end{subfigure}
  \caption{\textbf{Robustness to common video degradations.} Video AUROC (\%) of the
  main checkpoints, averaged over the seven test sets, as the strength of each
  degradation increases; the leftmost point is the unperturbed test set. Degradations
  are applied to every decoded frame: JPEG quality 80 to 10, Gaussian blur with
  $\sigma$ from 0.5 to 3 pixels, downscaling of the short side to 224, 112 and 64
  pixels, and Gaussian noise with $\sigma$ from 0.01 to 0.2 on pixel values in
  $[0,1]$. Per-dataset curves are given in
  ~\cref{fig:robust-auroc-jpeg,fig:robust-auroc-resize,fig:robust-auroc-blur,fig:robust-auroc-noise}, and AUROC, AP and EER per dataset in \cref{tab:robust-summary}.}
  \label{fig:robustness}
\end{figure*}

\subsubsection{Sensitivity to Initialization and Clip Length}

\paragraph{Random Seeds.}
We repeat training with five seeds to assess whether the comparison depends on a particular run (\cref{tab:ablation-training-seeds}). MARS achieves the highest mean AUROC across the seven
datasets at $90.69\pm1.02$\%, compared with $89.58\pm0.39$\% for DFD-HR, while GenD shows greater
variability at $66.26\pm5.62$\%. The variation is not uniform across datasets, as MARS has a standard deviation of 4.54\% on GenVideo but less than 1\% on FF++, CelebDF, DF-TIMIT, and UADFV. This indicates that its overall advantage persists across runs, although transfer to generated videos remains more sensitive to training randomness.

\paragraph{Number of Frames.}
Increasing clip length does not consistently improve detection (\cref{fig:ablation-training-frames}). MARS achieves 91.4--92.5\% average AUROC across 8--64 frames and leads at every tested clip length, whereas GenD peaks at 75.85\% with 16 frames before declining to 65.78\% with 64 frames. MARS therefore maintains strong performance with shorter clips under the tested training settings, although changing clip length also changes the number of training updates.

\subsubsection{Robustness to Processing}
We evaluate fixed checkpoints under compression, blur, downscaling, and noise to examine whether detection remains effective after video processing (\cref{fig:robustness}). MARS achieves the highest average AUROC at every tested JPEG quality, retaining 82.47\% at quality 40 compared with 77.55\% for DFD-HR. It also leads under downscaling to 112 pixels and blur up to $\sigma=1.5$, although ForAda performs better under stronger blur and downscaling.
Gaussian noise exposes a clearer weakness, with MARS dropping to 70.80\% at $\sigma=0.05$, below DFD-HR's 74.73\%. These results show that MARS preserves useful detection evidence under several processing conditions, but its advantage diminishes as degradation increases. DeMamba changes little across perturbations, yet remains near 50\% AUROC, illustrating that a small performance drop alone does not establish effective robustness.

\subsubsection{Faithfulness of Post-hoc Evidence}
\begin{figure}[t]
  \centering
  \begin{subfigure}[t]{0.49\linewidth}
    \centering
    \includegraphics[width=\linewidth]{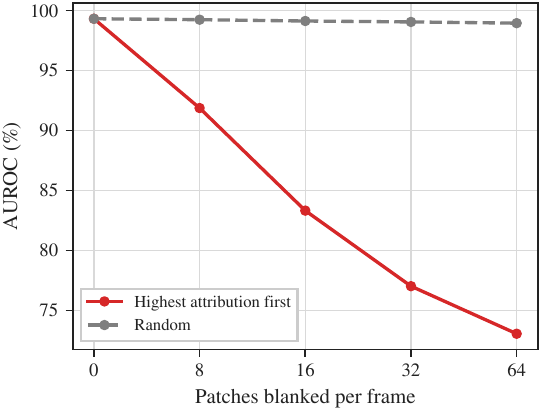}
    \caption{Patch deletion}
    \label{fig:faithfulness-patches}
  \end{subfigure}\hfill
  \begin{subfigure}[t]{0.49\linewidth}
    \centering
    \includegraphics[width=\linewidth]{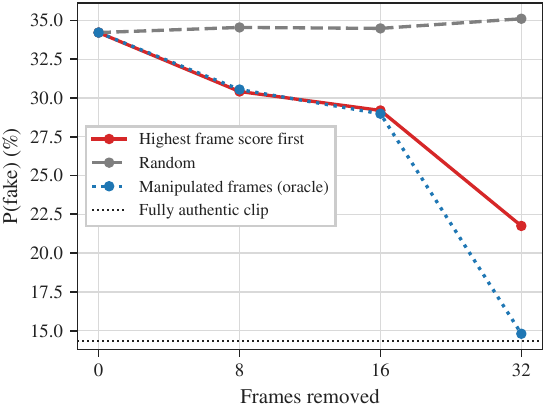}
    \caption{Frame removal}
    \label{fig:faithfulness-frames}
  \end{subfigure}
\caption{\textbf{Faithfulness of the MARS evidence.}
  (a)~AUROC, averaged over the seven test sets, after the $k$ highest-attribution
  patches or $k$ random patches are blanked in every frame. Each set contributes
  up to 100 manipulated clips that MARS detects and up to 100 authentic clips,
  so the unedited AUROC is close to 100.
  (b)~Probability of manipulation of FF++ clips containing a 32-frame manipulated
  segment, after the $k$ highest-scoring, random or manipulated (oracle) frames are
  removed; the dotted line marks the fully authentic clip.
  Per-dataset patch deletion and the 16-frame segment are given in
  \cref{fig:faith-patches,fig:faith-frames}, and the
  localization of manipulated frames in \cref{tab:localization}.}
  \label{fig:faithfulness}
\end{figure}

We test whether removing high-ranked regions or frames affects MARS's predictions more than random removal (\cref{fig:faithfulness}). On the selected evaluation clips, blanking the 64 highest-attribution patches per frame reduces average AUROC from 99.32\% to 73.05\%, whereas blanking the same number of random patches leaves it at 98.96\% (\cref{fig:faithfulness-patches}). The larger drop under attribution-guided removal supports the relevance of the spatial ranking to the detector's decisions.

For temporal evidence, we construct clips containing a 32-frame manipulated segment within authentic footage. Removing the 32 highest-scoring frames reduces the mean fake probability from 34.21\% to 21.75\%, while random removal leaves it at 35.10\% (\cref{fig:faithfulness-frames}). Removing the known manipulated frames lowers it further to 14.80\%, showing that the predicted ranking captures useful temporal evidence but does not fully recover the manipulated segment. Together, these interventions support the relevance of the displayed evidence to MARS's decisions, without establishing that every highlighted region is a forgery artifact.
 
\section{Conclusion}
\label{sec:conclusion}
We introduced MARS to investigate whether multiscale spatial and temporal feature differences complement adapted appearance representations for detecting facial manipulation and fully generated video. Under a shared protocol across seven datasets, MARS achieves the highest average AUROC among the compared detectors, while training on facial manipulations alone also transfers to generated videos. Ablations show that combining the paths improves facial and overall detection, with benefits that vary across datasets. Controlled removal tests further support the relevance of its signed patch and frame contributions to its predictions. These results support combining explicit feature comparisons with appearance representations for video forgery detection. However, MARS remains sensitive to Gaussian noise, strong blur, and downscaling, while its residual comparisons use fixed scales. Future work could explore adaptive scale and depth selection alongside training for greater robustness to video degradation. Extending MARS beyond visual evidence to incorporate audio for multimodal forgery detection.

\section*{Acknowledgments}
Resources used in preparing this research were provided, in part, by the Province of Ontario and the Government of Canada through CIFAR, as well as companies sponsoring the Vector Institute (\url{http://www.vectorinstitute.ai/\#partners}).
This research was funded by the European Union's Horizon Europe research and innovation programme under the AIXPERT project (Grant Agreement No.\ 101214389), which aims to develop an agentic, multi-layered, GenAI-powered framework for creating explainable, accountable, and transparent AI systems.

\section*{LLM Usage Statement}
We used LLMs for grammar and wording polish of the manuscript, and assisted coding. All technical content, experiments, and analysis were done by the authors.

{\small
\bibliographystyle{ieeenat_fullname}
\bibliography{main}
}

\appendix
\input{preprint_appendix}

\end{document}

%% file: preamble.tex
\usepackage[table]{xcolor}
\usepackage{pifont}
\usepackage{makecell}
\usepackage{algorithm}
\usepackage{algpseudocode}
\usepackage{xr-hyper}   

\algrenewcommand\alglinenumber[1]{\scriptsize #1:}

\makeatletter
\newcommand*{\addFileDependency}[1]{\typeout{(#1)}\@addtofilelist{#1}\IfFileExists{#1}{}{\typeout{No file #1.}}}
\makeatother

\definecolor{oursrow}{HTML}{E8EAF6}
\definecolor{deltarow}{HTML}{F5F5F5}
\newcommand{\cmark}{\ding{51}}
\newcommand{\xmark}{\ding{55}}

\newcommand{\grp}[2]{\multicolumn{#1}{l}{\textit{#2}}\\}
\newcommand{\pmstd}[2]{#1\textcolor{gray}{\,\scriptsize$\pm$#2}}

\newcommand{\FGdef}{F: mean AUROC over CelebDF, DF-TIMIT, UADFV, HiFiFace;
G: mean AUROC over GenVideo, DF26; Avg.: equal mean over the seven evaluation datasets (\%)}
\usepackage{subcaption}
\usepackage{multirow}
\usepackage{placeins}



%% file: preprint_appendix.tex
\maketitlesupplementary

\setcounter{table}{0}
\setcounter{figure}{0}
\setcounter{equation}{0}

\renewcommand{\thetable}{S\arabic{table}}
\renewcommand{\thefigure}{S\arabic{figure}}
\renewcommand{\theequation}{S\arabic{equation}}

\renewcommand{\theHtable}{supp.\arabic{table}}
\renewcommand{\theHfigure}{supp.\arabic{figure}}
\renewcommand{\theHequation}{supp.\arabic{equation}}


In this supplementary material, we provide detailed forward pass of our model, selected datasets, models configurations, additional experiments, and expanded detailed results. In~\cref{app:impl} we present additional details regarding MARS and visualization of the residuals streams, with a detailed steps how to extract evidences from the model.~\cref{app:datasets},~\cref{app:protocol} and~\cref{app:methods} present detailed setup about the selected datasets, followed protocol, and configurations of each method.~\cref{app:metrics} covers the selected metrics and how to read their results. In~\cref{app:additional-experiments} we provide additional experiments and ablations covering different backbone, and depth effect on residuals, training duration, different semantic-path frames, quantitative evidence and qualitative examples. Lastly~\cref{app:results} expands main tables and figures results to analyze per-dataset results and additional metrics.

\section{Additional MARS Implementation Details}
\label{app:impl}
 
This section complements \cref{sec:method} with the full forward pass, MARS hyperparameters, and visualizations of the residual streams.

\subsection{Forward Pass}
\label{app:forward}
 
Following our dual-path architecture, both paths use the same frozen attention and MLP weights of PE-Core-L but differ in depth and normalization parameters. The residual path passes every frame through blocks $1$ to $k$ once with the pretrained parameters $\gamma$, and all of its later operations act on the resulting patch tokens. The semantic path passes the $T_{\mathrm{sem}}$ selected frames through all $L$ blocks with its own parameters $\gamma^{\mathrm{sem}}$. Because $\gamma^{\mathrm{sem}}$ differs from $\gamma$ from the first block onward, these frames cannot reuse $Z_t$ and are encoded a second time. Besides the fake probability, the forward pass returns the patch weights $\alpha$, the per-frame semantic probabilities, the gate weight $w$, and the two standardized margins. We summarize the forward path in Algorithm~\ref{alg:mars}, and~\cref{tab:mars-hparams} lists the hyperparameters of MARS.

\begin{table}[t]
\centering\footnotesize
\setlength{\tabcolsep}{4pt}
\caption{\textbf{MARS hyperparameters.} Values shared with the other methods (optimizer, schedule, batch, epochs) are given in Appendix~\ref{app:protocol}.}
\resizebox{0.49\textwidth}{!}{%
\label{tab:mars-hparams}
\begin{tabular}{@{}ll@{}}
\toprule
Setting & Value\\
\midrule
\grp{2}{Encoder}
Backbone & PE-Core-L/14, $336\times336$, frozen\\
Patch grid, width & $24\times24$ ($N=576$), $D=1024$\\
Blocks, split block & $L=24$, $k=12$\\
\midrule
\grp{2}{Residual path}
Projection width & $d_r=256$\\
Scales & $\tau\in\{0.5,1,2,4\}$, $\sigma(\tau)=\sqrt{2\tau}$\\
Residual streams & 3 spatial, 3 temporal, 1 first difference\\
Fusion width & $d=1152$\\
Temporal transformer & 4 layers, 12 heads, MLP ratio 4, dropout 0.1, post-norm\\
Attention pool hidden width & 128 (tanh)\\
\midrule
\grp{2}{Semantic path and fusion}
Semantic frames & $T_{\mathrm{sem}}=16$\\
Tuned parameters & LayerNorm scale and shift (all 24 blocks), linear head\\
Margin standardization & running statistics, $\epsilon=10^{-5}$, momentum $0.1$\\
Gate initialization & $g=0$, $b_g=0$ ($w=0.5$)\\
Loss weights & semantic : residual : fused $=1:1:1$\\
\midrule
Trainable parameters & 67.54M (residual path 67.43M, semantic path 0.11M)\\
\bottomrule
\end{tabular}
}
\end{table}

\begin{algorithm}[t]
\caption{MARS forward pass for one clip.}
\label{alg:mars}
\small
\begin{algorithmic}[1]
\Require Clip $X=(x_1,\ldots,x_T)$, validity mask $m$, residual depth $k$
\Ensure Fake probability $P(\mathrm{fake}\mid X)$ and diagnostic outputs

\Statex \textit{// Frozen intermediate features for the residual path}
\State $T_v\gets\sum_t m_t$
\State $Z_t\gets\mathcal{E}_{1:k}(x_t;\gamma)$ for $t=1,\ldots,T$ \Comment{\cref{eq:split}}
\State $F\gets$ patch tokens of $Z_{1:T}$

\Statex \textit{// Residual path: spatial and temporal differences}
\State $\bar F\gets\mathrm{LN}(F)$;\ \ $R\gets\bar F W_r+b_r$ \Comment{\cref{eq:residual_projection}}
\For{$j=1,\ldots,J-1$}
    \State $r_j^{\mathrm{sp}}\gets\mathcal{S}_{\tau_j}R-\mathcal{S}_{\tau_{j+1}}R$ \Comment{\cref{eq:bands}}
    \State $r_j^{\mathrm{tm}}\gets\mathcal{T}_{\tau_j}^{m}R-\mathcal{T}_{\tau_{j+1}}^{m}R$
\EndFor
\State Compute $r^{\Delta}$ \Comment{\cref{eq:diff}}
\State Combine appearance and residual streams into $U$ \Comment{\cref{eq:fusion}}
\State Add positional encodings to $U$ \Comment{\cref{eq:position}}

\Statex \textit{// Residual path: temporal modeling and readout}
\State $u_n\gets T_v^{-1}\sum_t m_t\,\mathrm{TT}(U_{\cdot,n};m)_t$ for $n=1,\ldots,N$ \Comment{\cref{eq:tpool}}
\State Compute $\alpha$, $z_{\mathrm{res}}$, and $\ell_{\mathrm{res}}$ \Comment{\cref{eq:spool}}

\Statex \textit{// Semantic path: adapted encoder on selected frames}
\State Select $t_1,\ldots,t_{T_{\mathrm{sem}}}$ \Comment{\cref{eq:semantic_sampling}}
\For{$i=1,\ldots,T_{\mathrm{sem}}$}
    \State $q_{t_i}\gets\mathcal{E}_{1:L}(x_{t_i};\gamma^{\mathrm{sem}})$ \Comment{\cref{eq:semantic_encoder}}
    \State $\hat q_i\gets q_{t_i}/\|q_{t_i}\|_2$
\EndFor
\State Compute $\ell_{\mathrm{sem}}$ and $e_{\mathrm{sem}}$ \Comment{\cref{eq:semantic}}

\Statex \textit{// Adaptive fusion}
\State Standardize $\delta_{\mathrm{sem}}$ and $\delta_{\mathrm{res}}$ \Comment{\cref{eq:standardize}}
\State Compute $w$ and $\delta$ \Comment{\cref{eq:gate}}
\State \Return $\operatorname{sigmoid}(\delta)$, $\alpha$, $\{p_i^{\mathrm{sem}}\}$, $w$, $\tilde\delta_{\mathrm{sem}}$, $\tilde\delta_{\mathrm{res}}$
\end{algorithmic}
\end{algorithm}

\subsection{Residuals Visualization}
\label{app:residual-visualization}

\begin{figure*}[t]
  \centering
  \includegraphics[width=\textwidth]{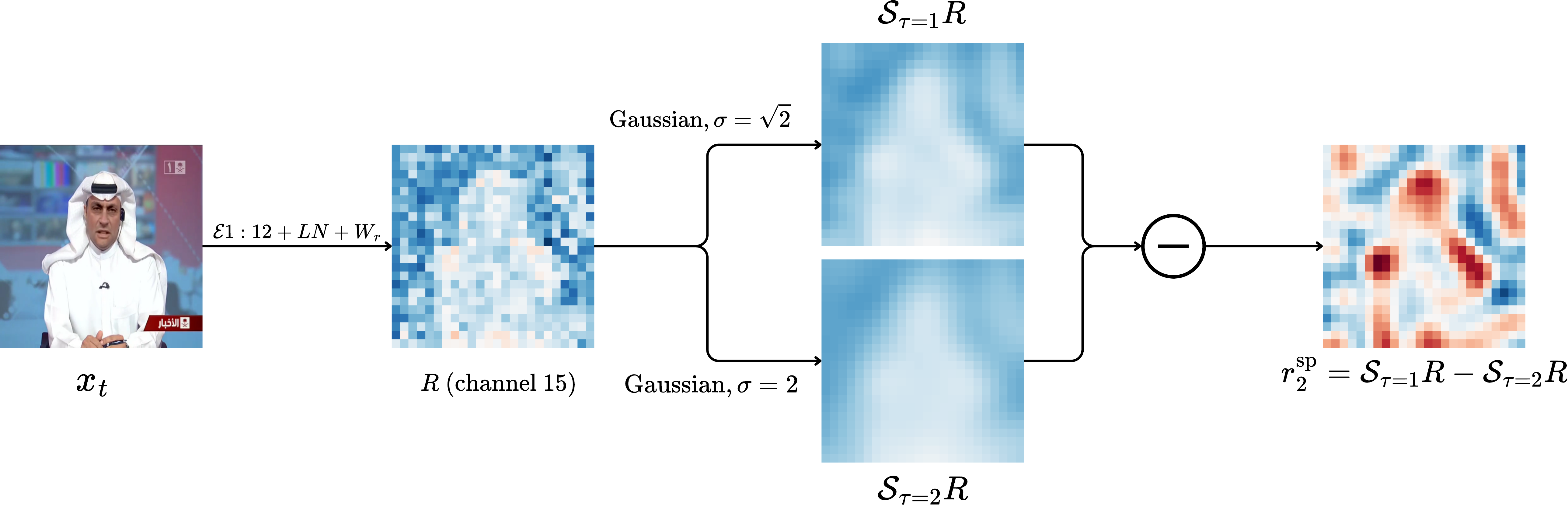}
\caption{\textbf{Spatial comparison on trained MARS features.} A frame of an FF++ Deepfakes video is encoded by blocks 1--12 of the frozen encoder ($\mathcal{E}_{1:12}$, \cref{eq:split}), then normalized and projected to $R$ (\cref{eq:residual_projection}). We show channel~15 of $R$, its smoothings at $\tau=1$ and $\tau=2$, and their difference $r_2^{\mathrm{sp}}$ (\cref{eq:bands}). Maps of $R$ and its smoothings share one zero-centered color scale; $r_2^{\mathrm{sp}}$ uses its own.}
  \label{fig:spatial_residual}
\end{figure*}

\begin{figure}[t]
    \centering
    \includegraphics[width=0.49\columnwidth]{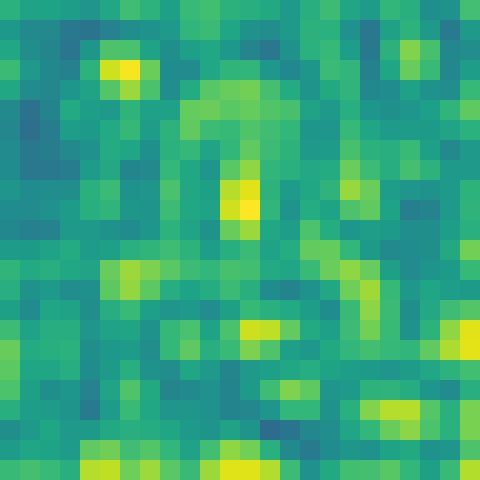}\hfill
    \includegraphics[width=0.49\columnwidth]{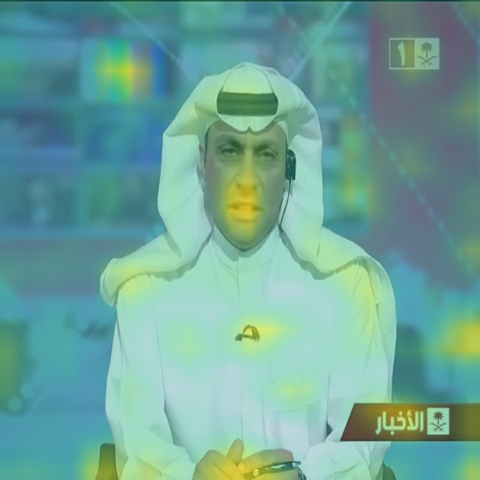}
    \caption{Spatial residual energy of the frame in
   ~\cref{fig:spatial_residual}. For each patch $n$, the energy is
    $\|r_{2,n}^{\mathrm{sp}}\|_2$, the norm of the band $r_2^{\mathrm{sp}}$
    ($\tau=1$ vs.\ $2$) over all $d_r=256$ channels.
    Left: on the $24\times24$ patch grid.
    Right: overlaid on the frame.
    Energy measures response magnitude, not where the detector looks.}
    \label{fig:spatial_energy}
\end{figure}

We visualize the residual streams of \cref{eq:bands} using features from one FF++ Deepfakes video processed by the trained MARS checkpoint.~\cref{fig:spatial_residual} follows one projected channel of $R$ through spatial smoothing at $\tau=1$ and $\tau=2$, then subtracts the two outputs to obtain $r_2^{\mathrm{sp}}$. The larger scale smooths over a wider neighborhood, so their difference emphasizes variation between the two scales while suppressing content preserved by both. \cref{fig:spatial_energy} summarizes this response using the $\ell_2$ norm across all $d_r$ channels at each patch. This energy measures the strength of the residual response, which can also arise from natural texture, rather than its contribution to the detector's decision.

\begin{figure}[t]
    \centering
    \includegraphics[width=1.0\columnwidth]{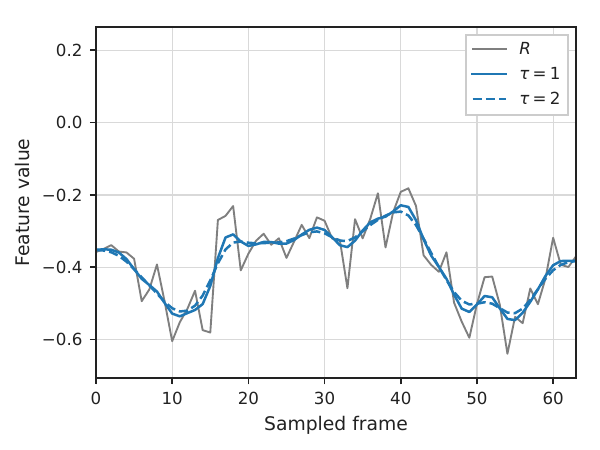}
    \par\vspace{2pt}
    \includegraphics[width=1.0\columnwidth]{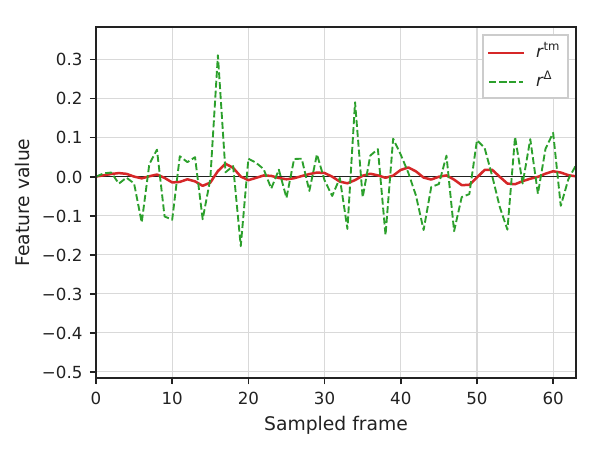}
    \caption{Temporal comparisons at a fixed patch and channel
    in an FF++ Deepfakes clip.
    Top: channel~15 of $R$, the PE block-12 features after projection
    (\cref{eq:residual_projection}), over the 64 frames, with its
    Gaussian smoothings at $\tau=1$ and $\tau=2$.
    Bottom: their difference $r_2^{\mathrm{tm}}$ and the
    consecutive-frame difference $r^{\Delta}$.
    The patch is selected by maximum interior attention,
    and the channel by the largest spatial variance of the band ($\tau=1$ vs.\ $2$) over interior patches of the fake frame.}
    \label{fig:temporal_comparisons}
\end{figure}

\cref{fig:temporal_comparisons} shows the temporal comparisons at one patch position and feature channel. Smoothing along the frame axis reduces short fluctuations, and subtracting consecutive smoothing scales produces $r_2^{\mathrm{tm}}$. The first difference $r^\Delta$ (\cref{eq:diff}) instead compares consecutive sampled frames, capturing changes at the shortest observed interval.


\subsection{Spatial and Temporal Evidence}
\label{app:evidence}

We compute signed patch and frame contributions from the
main MARS checkpoint to inspect which regions and frames
support its prediction.

Let $h_{t,n}$ be the temporal transformer's output for frame $t$ and patch $n$. The contribution of this feature to the residual path's fake-versus-real margin is

\begin{equation}
c_{t,n}
=
\frac{m_t\alpha_n}{T_v}
\omega_{\mathrm{res}}^\top h_{t,n},
\label{eq:residual_contribution}
\end{equation}

where $\omega_{\mathrm{res}}$ is the difference between the classifier's fake and real weight vectors. Summing $c_{t,n}$ over frames and patches, then adding the classifier's bias difference, recovers the residual margin.

For each frame selected by the semantic path, we also retain its fake-versus-real margin,
\begin{equation}
\delta_i^{\mathrm{sem}}
=
\phi_{\mathrm{sem}}(\hat q_i)^{(1)}
-
\phi_{\mathrm{sem}}(\hat q_i)^{(0)}.
\label{eq:semantic_frame_margin}
\end{equation}

We combine these contributions using the same gate weight and normalization as the detector,

\begin{equation}
\begin{aligned}
s_t ={}&
\frac{1-w}{\sqrt{v_{\mathrm{res}}+\epsilon}}
\sum_n c_{t,n}\\
&+
\frac{w}{T_{\mathrm{sem}}\sqrt{v_{\mathrm{sem}}+\epsilon}}
\sum_{i:t_i=t}\delta_i^{\mathrm{sem}}.
\end{aligned}
\label{eq:frame_scores}
\end{equation}

A higher $s_t$ means that frame $t$ contributes more toward a fake prediction. Frames not selected by the semantic path receive only the residual contribution. The patch map uses $(1-w)c_{t,n}/\sqrt{v_{\mathrm{res}}+\epsilon}$, summed over time to rank spatial positions.

These scores include the classifier's signed contributions, rather than attention weights alone. Their sum, together with the bias and normalization offsets, recovers the
fused margin to numerical precision; the reported float32 error is approximately $5\times10^{-7}$. The tests below examine whether these scores also identify input regions and frames that affect the prediction.

\section{Dataset Details}
\label{app:datasets}

\paragraph{FaceForensics++.}
FaceForensics++~\cite{rossler2019faceforensics++} provides
authentic videos and facial manipulations produced by Deepfakes,
Face2Face, FaceSwap, and NeuralTextures. We use c23 compression and the official splits, comprising 3,600 training videos, 700 validation videos, and 700 test videos. The training split contains 720 authentic videos and 720 videos from each manipulation method, while the validation and test splits each contain 140 authentic videos and 140 videos per method. The validation split is used for checkpoint selection. The videos are YouTube recordings of mostly frontal faces, typically around 480p. The median video has 414 frames at 25\,fps (about 17\,s) in the training split, 398 frames at 30\,fps (about 13\,s) in the validation split, and 444 frames at 25\,fps (about 18\,s) in the test split.

\paragraph{GTA-V.}
We use rendered sequences from SAIL-VOS~3D~\cite{hu2021sail} as fully synthetic videos. The training split contains 141 sequences with a median of 336 frames, and the held-out evaluation split contains 30 sequences with a median of 389 frames, all at
$1280\times800$ resolution.

\paragraph{CelebDF-V2.}
CelebDF-V2~\cite{li2020celeb} contains face swaps generated
from celebrity interviews. We use its official test list of 518 videos, comprising 340 manipulated videos and 178 authentic videos.
The authentic pool includes 108 celebrity interview videos
and 70 YouTube videos. The evaluated videos are typically around 480p,
at 30\,fps, with a median of 372 frames (approximately 12\,s).

\paragraph{DeepFakeTIMIT.}
DeepFakeTIMIT~\cite{korshunov2018deepfakesdftmit}
contains face swaps at two quality levels, with 320 videos
at each level. We pair these 640 manipulated videos with the 320 corresponding VidTIMIT~\cite{sanderson2009multi} recordings, giving 960 evaluation videos. The authentic recordings cover 32 subjects reading ten sentences each, and matching uses the subject directory and sentence identifier rather than the inserted identity.
Each authentic recording is stored as a sequence of numbered
JPEG frames, which we read in numerical order as one video.
The manipulated videos have a resolution of $512\times384$ and a frame rate of 25\,fps. All videos have a median of 103 frames (about 4\,s).

\paragraph{HiFiFace.}
HiFiFace~\cite{ijcai2021-157-hififace} releases 1,000 face-swapped videos generated from FaceForensics++ footage, without an accompanying authentic-video set. We identify the target video from each filename and retain only swaps whose target index belongs to the official FaceForensics++ test split, excluding targets from the training and validation splits. 
Each retained swap is paired with its corresponding c23 authentic target video, giving a balanced evaluation set of 140 authentic and 140 manipulated videos that share the underlying footage. The videos are typically around 480p at 25\,fps, with a median of 460 frames (about 18\,s).

\paragraph{UADFV.}
UADFV~\cite{yang2019exposing} contains 49 authentic videos
and their 49 face-swapped counterparts. We evaluate all 98 videos. Their resolutions reach approximately 500 pixels on the short side, with a frame rate of 30\,fps and a median of 204 frames (about 7\,s).

\paragraph{GenVideo.}
We use the provided validation split of GenVideo~\cite{chen2026demamba} for evaluation,
containing 18,286 videos in the evaluated manifest. These comprise 10,000 authentic videos and 8,286 generated videos across ten groups: Crafter, Gen2, HotShot, Lavie, ModelScope, MoonValley, MorphStudio, Show-1, Sora, and WildScrape. Group sizes range from 56 videos for Sora to 1,400 for Lavie. We retain the provided authentic-video pool without constructing additional real--fake pairings. The authentic videos have a median of 41 frames at 3\,fps (about 14\,s). For generated videos, the per-generator median ranges from 8 frames (HotShot) to 450 (Sora), at 8--50\,fps. Most generated videos last 1--4\,s, while the Sora videos are approximately 15\,s long. Although frame rate can act as a shortcut in GenVideo, MARS’s semantic path scores frames independently of their spacing, samples only from genuine frames, and still reaches 93.59\% AUROC on GenVideo by itself (\cref{tab:ap-ablation-training}), comparable to full MARS (93.44\%), so MARS’s transfer does not require this cue.

\paragraph{DF26.}
DF26~\cite{shykula2026df26} contributes 2,691 videos,
comprising 271 authentic recordings and 2,420 generated videos.
The authentic recordings show people speaking in three settings: direct-to-camera recordings (100), official statements (71), and studio interviews (100). The generated videos cover ten generator or generation-mode groups from seven model families: Grok Imagine, HunyuanVideo~1.5 (T2V and I2V), Kling~3.0, LTX~2.3 (T2V and I2V), Veo~3.1, Wan~2.2 (T2V and I2V), and Wan~2.6. Each group contains between 197 and 271 videos. We use the provided evaluation set and authentic-video pool without constructing additional pairings. All videos have a resolution of $1280\times720$ and last 4--6\,s, with 81 to 150 frames at 16--30\,fps.

For GenVideo and DF26, per-generator evaluation compares
each group's generated videos against all authentic videos
from the same dataset. Pooled evaluation includes each authentic and generated video once.

\begin{table*}[t]
\centering
\scriptsize
\setlength{\tabcolsep}{3.5pt}
\renewcommand{\arraystretch}{1.12}
\caption{\textbf{Summary of the datasets used for training and evaluation.} Video counts are reported for each selected split, while resolution, frame rate, duration, and frame count describe the original videos before preprocessing (frame counts are medians). For GenVideo, values are given separately for real and fake videos; fake durations describe most generated videos, and fake frame counts are the range of per-generator medians. -- indicates unreported measurements.}
\label{tab:datasets}
\resizebox{\textwidth}{!}{%
\begin{tabular}{@{}lllrrcccc@{}}
\toprule
Dataset & Split / use & Fake content &
Real & Fake &
Resolution & FPS & Duration (s) & Median frames\\
\midrule
FF++ c23
& Train & Facial manipulation
& 720 & 2,880
& $\sim$480p & 25 & $\sim$17 & 414\\
FF++ c23
& Validation & Facial manipulation
& 140 & 560
& $\sim$480p & 30 & $\sim$13 & 398\\
FF++ c23
& Test & Facial manipulation
& 140 & 560
& $\sim$480p & 25 & $\sim$18 & 444\\
GTA-V
& Train & Rendered scenes
& 0 & 141
& $1280\times800$ & -- & -- & 336\\
GTA-V
& Test & Rendered scenes
& 0 & 30
& $1280\times800$ & -- & -- & 389\\
\midrule
CelebDF-V2
& Test & Facial manipulation
& 178 & 340
& $\sim$480p & 30 & $\sim$12 & 372\\
DeepFakeTIMIT
& Test & Facial manipulation
& 320 & 640
& $512\times384$ (fake) & 25 & $\sim$4 & 103\\
HiFiFace
& Test & Facial manipulation
& 140 & 140
& $\sim$480p & 25 & $\sim$18 & 460\\
UADFV
& Test & Facial manipulation
& 49 & 49
& Short side $\lesssim500$ & 30 & $\sim$7 & 204\\
GenVideo
& Validation used as test & Generated video
& 10,000 & 8,286
& Variable & 3 (real); 8--50 (fake) & $\sim$14 (real); 1--4 (fake) & 41 (real); 8--450 (fake)\\
DF26
& Test & Generated video
& 271 & 2,420
& $1280\times720$ & 16--30 & 4--6 & 81--150\\
\bottomrule
\end{tabular}%
}
\end{table*}

\section{Shared Training and Evaluation Protocol}
\label{app:protocol}

All methods, including ours, are trained and evaluated within a single pipeline, so that they differ only in their networks and training objectives.

\paragraph{Input.}
Each video is decoded in full, and every second frame is retained. Following UNITE~\cite{kundu2025towards}, the retained frames are split into consecutive, non-overlapping clips of 64 frames, each treated as one sample that carries the label of its source video. A clip with fewer than 64 frames, which occurs at the end of a video or for videos shorter than one clip, is padded by repeating its last frame, and a mask marking the padded frames is provided to the network. Frames are used in full, without face cropping or data augmentation, and are resized to each method's input resolution, scaled to $[-1,1]$, and converted by its encoder wrapper.

\paragraph{Training and model selection.}
Unless stated otherwise, every method is trained on the FaceForensics++ c23 training split together with the GTA-V training sequences, while the FaceForensics++ validation split is used for model selection. Optimization uses AdamW~\cite{loshchilov2017decoupled} with a learning rate of $10^{-4}$, a weight decay of $10^{-4}$, and $(\beta_1, \beta_2) = (0.9, 0.999)$, halving the learning rate every 1,000 optimizer steps. Each step uses an effective batch of 32 clips, with gradient accumulation for methods whose memory footprint requires smaller micro-batches, and gradients are clipped to a norm of 1.0. Training runs for three epochs in bfloat16 mixed precision on four NVIDIA A40 GPUs with distributed data parallelism. After each epoch, the model is evaluated on the FaceForensics++ validation videos, and the epoch with the highest video-level AUROC is retained. The main results use a fixed seed, which fixes both the data order and the initialization of all trainable parameters.

\paragraph{Testing.}
At test time, every clip of a video is scored, and the video score is the mean of its clip probabilities of being fake, so that each video contributes once regardless of its length. AUROC, AP, and EER are then computed over the videos of each dataset.

\section{Compared Methods}
\label{app:methods}

Each compared method is implemented within the shared pipeline of \cref{app:protocol}: we use its network and training objective from the original paper and released code, except where noted, while inputs, optimization, model selection, and evaluation follow the shared protocol. For methods that classify individual frames, the clip logit is the mean of the frame logits.~\cref{tab:methods} summarizes the configuration of each method.

\paragraph{MARS (ours).}
MARS splits a single frozen PE-Core-L encoder~\cite{bolya2026perception} at $336\times336$ after block 12 into the residual and semantic paths described in~\cref{sec:method}. The residual path processes all 64 frames of a clip, whereas the semantic path passes 16 frames through the whole encoder with its own trainable copies of the LayerNorm parameters, and a gate combines the scores of the two paths. The model is trained with a cross-entropy loss on each path and on the fused prediction.

\paragraph{UNITE.}
UNITE~\cite{kundu2025towards} encodes each frame with a frozen SigLIP-so400m encoder~\cite{zhai2023sigmoid} at $384\times384$ and models time with a four-block per-token temporal transformer over the 64 frames of a clip, followed by attention pooling over tokens and
a linear classifier. It is trained with cross-entropy and the attention-diversity loss, each weighted by 0.5.

\paragraph{GenD.}
GenD~\cite{yermakov2026deepfake} tunes only the normalization layers of the PE-Core-L encoder~\cite{bolya2026perception} at $336\times336$ and classifies the L2-normalized pooled feature of each frame with a linear layer. It is trained with cross-entropy together with the
uniformity and label-aware alignment losses on the unit sphere, weighted by 0.5 and 0.1, respectively.

\paragraph{ForAda.}
ForAda~\cite{cui2025forensics} steers a frozen CLIP ViT-L/14~\cite{radford2021learning} at $224\times224$ with a side adapter, built from the first eight blocks of a ViT-Tiny that reads the frame at $256\times256$. The adapter produces attention biases through which 128 learnable query tokens attend to the CLIP patch tokens, and the resulting queries are mapped to the frame logit, with the whole model trained by cross-entropy.

\paragraph{DFD-HR.}
DFD-HR~\cite{sun2026dfd} views each frame at $448\times448$ as one global $224\times224$ view and a $2\times2$ grid of $224\times224$ tiles, which pass through a frozen CLIP ViT-L/14~\cite{radford2021learning} equipped with four-expert mixture-of-experts adapters after the attention and MLP of every block. In blocks 21--23, a layer router allows a view to exit early and a token router forwards the top 75\% of tokens, after which the global class token and a query-attention summary of the four tile class tokens are classified jointly. The model is trained with cross-entropy and the token-ranking loss, the latter weighted by 0.1.

\paragraph{DeMamba.}
DeMamba~\cite{chen2026demamba} fully fine-tunes an X-CLIP B/16 encoder~\cite{ni2022expanding} at $224\times224$, which exchanges messages among groups of eight consecutive frames. Its global
branch averages the class tokens over the clip, while its detail branch reads the tokens of each spatial zone across frames with a Mamba block, and the two branches are combined into one logit trained with binary cross-entropy.

\begin{table*}[t]
\centering
\scriptsize
\setlength{\tabcolsep}{3.5pt}
\renewcommand{\arraystretch}{1.12}
\caption{\textbf{Configuration of the compared methods.} All methods follow the shared protocol of
Appendix~\ref{app:protocol}, with an effective batch of 32 clips of 64 frames. Input is the
resolution read by the encoder.}
\label{tab:methods}
\resizebox{\textwidth}{!}{%
\begin{tabular}{@{}lllclcr@{}}
\toprule
Method & Encoder & Trained components & Input & Clip aggregation & Objective &
Trainable\\
\midrule
UNITE & SigLIP-so400m (frozen) & Temporal transformer, head & 384 & Temporal transformer & CE + AD & 63.91M\\
GenD & PE-Core-L & Normalization layers, head & 336 & Mean of frame logits & CE + uniformity + alignment & 0.11M\\
ForAda & CLIP ViT-L/14 (frozen) & Side adapter, queries, head & 224 / 256 & Mean of frame logits & CE (mask-based terms omitted) & 5.98M\\
DFD-HR & CLIP ViT-L/14 (frozen) & MoE adapters, token routers, head & 448 & Mean of frame logits & CE + ranking & 103.48M\\
DeMamba & X-CLIP B/16 & Full encoder, Mamba, head & 224 & Clip-level & BCE & 125.70M\\
Effort & CLIP ViT-L/14 (frozen) & Rank-1 SVD residuals, head & 224 & Mean of frame logits & CE + orthogonality & 0.20M\\
ReStraV & DINOv2 ViT-S/14 (frozen) & MLP & 224 & Trajectory statistics & BCE & 3.5K\\
\midrule
MARS (ours) & PE-Core-L (frozen) & Residual path, semantic LayerNorms, heads, gate & 336 & Two-path gate & CE (per path + fused) & 67.54M\\
\bottomrule
\end{tabular}%
}
\end{table*}

\paragraph{Effort.}
Effort~\cite{yan2024effort} applies singular value decomposition to the query, key, value, and output projections of every block of a frozen CLIP ViT-L/14~\cite{radford2021learning} at $224\times224$, freezing the principal singular directions and training a rank-one residual, and classifies the class token of each frame with a linear layer. It is trained with cross-entropy together with the orthogonality and singular-value constraints of the paper, weighted by 1.0.

\paragraph{ReStraV.}
ReStraV~\cite{interno2026ai} encodes each frame with a frozen DINOv2
ViT-S/14~\cite{oquab2023dinov2} at $224\times224$ and treats the normalized tokens of successive frames as a trajectory. Twenty-one statistics of the step lengths and turning angles along this trajectory are standardized and classified by a two-layer MLP trained with binary cross-entropy.

\section{Evaluation Metrics}
\label{app:metrics}

All metrics are computed at the video level and reported as percentages, with fake videos as the positive class.

\paragraph{Video scores.}
A video $V$ divided into clips $X_1,\ldots,X_C$ receives the score
\begin{equation}
s_V = P(\mathrm{fake}\mid V) = \frac{1}{C}\sum_{c=1}^{C} P(\mathrm{fake}\mid X_c) \in [0,1].
\label{eq:video_score}
\end{equation}
For an evaluation set, $\mathcal{V}_0$ and $\mathcal{V}_1$ denote its authentic ($y=0$) and fake ($y=1$) videos. For a threshold $\theta\in[0,1]$, a video is predicted fake when $s_V\ge\theta$, giving the false-positive and false-negative rates

\begin{equation}
\begin{aligned}
\mathrm{FPR}(\theta)&=\frac{|\{V\in\mathcal{V}_0 : s_V\ge\theta\}|}{|\mathcal{V}_0|},\\
\mathrm{FNR}(\theta)&=\frac{|\{V\in\mathcal{V}_1 : s_V<\theta\}|}{|\mathcal{V}_1|}.
\end{aligned}
\label{eq:rates}
\end{equation}

\paragraph{AUROC.}
The area under the ROC curve equals the probability that a randomly drawn fake video scores higher than a randomly drawn authentic video, with ties counted as one half,

\begin{equation}
\begin{aligned}
\mathrm{AUROC}=\frac{1}{|\mathcal{V}_0|\,|\mathcal{V}_1|}
&\sum_{V\in\mathcal{V}_1}\sum_{V'\in\mathcal{V}_0}
\Big(\mathbb{1}[s_V>s_{V'}]\\
&+\tfrac{1}{2}\,\mathbb{1}[s_V=s_{V'}]\Big).
\end{aligned}
\label{eq:auroc}
\end{equation}

AUROC depends only on how the scores rank the videos, so it requires no threshold and is unaffected by any monotonic rescaling of the scores.

\paragraph{EER.}
The equal error rate corresponds to the threshold where the false-positive and false-negative rates defined in \cref{eq:rates} coincide. On the finite evaluation set, we approximate it at the threshold where they are closest,

\begin{equation}
\begin{aligned}
\mathrm{EER}&=\tfrac{1}{2}\big(\mathrm{FPR}(\theta^\ast)+\mathrm{FNR}(\theta^\ast)\big),\\
\theta^\ast&=\arg\min_{\theta}\,\big|\mathrm{FPR}(\theta)-\mathrm{FNR}(\theta)\big|.
\end{aligned}
\label{eq:eer}
\end{equation}

Because $\theta^\ast$ is chosen on the evaluated set itself, EER describes an operating point that balances the two error rates rather than performance at a threshold fixed in advance.

\paragraph{AP.}
Let $\theta_1>\theta_2>\cdots$ be the distinct video scores of the set, and let $\mathrm{Prec}_j$ and $\mathrm{Rec}_j$ be the precision and recall of the fake class at threshold $\theta_j$, with $\mathrm{Rec}_0=0$. Average precision is
\begin{equation}
\mathrm{AP}=\sum_{j}\big(\mathrm{Rec}_j-\mathrm{Rec}_{j-1}\big)\,\mathrm{Prec}_j .
\label{eq:ap}
\end{equation}
Unlike AUROC, AP depends on the proportion of fake videos in the set, so its values are comparable across methods on the same set but not across sets with different class balance.

\paragraph{Per-generator, macro, and pooled results.}
For GenVideo and DF26, let $\mathcal{G}$ be the set of generators and $\mathcal{V}_1^{(g)}$ the fake videos of generator $g$. The per-generator metric compares $\mathcal{V}_1^{(g)}$ with the full authentic pool $\mathcal{V}_0$ of the same dataset, so all generators share one set of authentic videos. The macro value averages these per-generator metrics with equal weight,
\begin{equation}
\mathrm{AUROC}_{\mathrm{macro}}=\frac{1}{|\mathcal{G}|}\sum_{g\in\mathcal{G}}\mathrm{AUROC}\big(\mathcal{V}_0,\mathcal{V}_1^{(g)}\big),
\label{eq:macro}
\end{equation}
and analogously for EER. The pooled value is computed once on $\mathcal{V}_0\cup\bigcup_{g}\mathcal{V}_1^{(g)}$, so each video counts once and generators with more videos carry more weight.

\section{Additional Experiments}
\label{app:additional-experiments}

\subsection{Residual Extraction Depth and Backbone}
We move the split of the residual path from block 12 to blocks 6, 18, and 24 (~\cref{tab:ap-split-backbone}). Block 12 gives the best generated-video mean and seven-dataset average, whereas later splits at blocks 18 and 24 improve facial manipulation by about 1\% but lower generated video by up to 8\%. An earlier split at block 6 lowers both. This suggests that cues of generated video are best exposed in intermediate features, which still retain local detail, while deeper features become more semantic and favor facial manipulation, consistent with the observation that the strongest PE features lie inside the network rather than at its output~\cite{bolya2026perception}.

We also replace PE-Core-L with SigLIP-so400m~\cite{zhai2023sigmoid}, trained with image--text contrastive learning, and DINOv3-L~\cite{simeoni2025dinov3}, trained with self-supervision on images for dense prediction. Both keep facial manipulation within about 5\% of PE but drop generated video by more than 40\%, below chance. 
One possible contributor is that PE-Core is also trained on video, whereas SigLIP and DINOv3 are image-only, although other differences in data and objectives remain uncontrolled. Overall, we observe that the residual design transfers to facial manipulation across backbones, but its generated-video performance depends on the pretraining of the encoder.


\begin{table}[t]
\centering\small
\setlength{\tabcolsep}{6pt}
\caption{\textbf{Encoder depth and backbone ablation.}
Results with a fixed seed and three epochs. Each row changes
only the listed setting from the highlighted default
(PE-Core-L, block 12). \FGdef. Best results are \textbf{bold} and second-best results
are \underline{underlined} in each column.}
\label{tab:ap-split-backbone}
\begin{tabular}{@{}llccc@{}}
\toprule
Encoder & Split block & F & G & Avg.\\
\midrule
\grp{5}{Split layer}
PE & 6 & 89.97 & \underline{81.04} & 88.58\\
\rowcolor{oursrow}
PE & \textbf{12} & 92.95 & \textbf{84.63} & \textbf{91.43}\\
PE & 18 & \textbf{94.06} & 77.04 & 89.89\\
PE & 24 & \underline{93.61} & 80.80 & \underline{90.68}\\
\midrule
\grp{5}{Backbone (split at block 12)}
SigLIP-so400m & 12 & 88.26 & 42.47 & 76.60\\
DINOv3-L & 12 & 88.31 & 40.34 & 75.88\\
\bottomrule
\end{tabular}
\end{table}
\begin{table}[t]
\centering\footnotesize
\setlength{\tabcolsep}{3pt}
\caption{\textbf{Training duration across methods.}
Facial (F) and generated-video (G) mean AUROC (\%),
using 64-frame clips and a fixed seed.
Best results are \textbf{bold} and second-best results
are \underline{underlined} in each column.}
\label{tab:training-fg}
\resizebox{\linewidth}{!}{%
\begin{tabular}{@{}l*{6}{c}@{}}
\toprule
& \multicolumn{2}{c}{3 epochs}
& \multicolumn{2}{c}{5 epochs}
& \multicolumn{2}{c}{10 epochs}\\
\cmidrule(lr){2-3}
\cmidrule(lr){4-5}
\cmidrule(lr){6-7}
Method & F & G & F & G & F & G\\
\midrule
\multicolumn{7}{l}{\textit{Frame-level}}\\
ForAda
& 87.72 & 52.37 & 85.76 & 53.31 & 84.79 & 53.14\\
Effort
& 73.01 & 47.21 & 74.26 & 47.16 & 75.23 & 48.05\\
GenD
& 66.35 & 62.27 & 69.40 & \underline{77.67}
& 88.15 & \underline{84.96}\\
DFD-HR
& \textbf{94.00} & \underline{77.21}
& \textbf{94.81} & 77.29
& \textbf{94.78} & 75.58\\
\midrule
\multicolumn{7}{l}{\textit{Clip-level}}\\
UNITE
& 81.34 & 54.71 & 85.09 & 49.84 & 86.93 & 49.22\\
DeMamba
& 49.70 & 51.30 & 48.27 & 49.78 & 48.15 & 49.60\\
ReStraV
& 39.78 & 34.86 & 41.73 & 33.07 & 43.91 & 31.57\\
\midrule
\rowcolor{oursrow}
\textbf{MARS (Ours)}
& \underline{92.95} & \textbf{84.63}
& \underline{93.91} & \textbf{83.32}
& \underline{92.86} & \textbf{85.03}\\
\bottomrule
\end{tabular}}
\end{table}
\subsection{Training Duration}
We compare training for three, five, and ten epochs while
keeping the clip length at 64 frames
(\cref{tab:training-fg}). GenD benefits most from longer
training, reaching 88.15\% facial and 84.96\% generated-video
AUROC at ten epochs, compared with 92.86\% and 85.03\%
for MARS. MARS and DFD-HR change less across these budgets:
MARS's overall AUROC varies from 91.43\% to 91.70\%,
while DFD-HR retains the highest facial AUROC.
Longer training therefore narrows GenD's gap on generated
videos, while MARS maintains stronger facial performance.

\subsection{Fairness of Comparisons}
We examine whether the shared training settings explain the lower results of DeMamba and GenD (\cref{tab:ap-fairness}). For DeMamba, which fine-tunes the full encoder, we reduce the learning rate from $10^{-4}$ to $10^{-5}$. Average AUROC increases from 53.29\% at three epochs to 65.14\% at ten, but the improvement is concentrated on facial datasets, while GenVideo falls from 60.88\% to 28.51\%.
For GenD, removing the alignment and uniformity losses raises three-epoch average AUROC from 65.78\% to 90.62\%, close to the semantic-only variant's 90.24\%. These results show that the shared budget and training objectives help explain the baseline gaps.

We also evaluate the semantic-only variant across five seeds, obtaining $87.40\pm3.93\%$ average AUROC compared with $90.69\pm1.02\%$ for full MARS (\cref{tab:ablation-training-seeds}). Combining the paths raises mean facial AUROC from 86.24\% to 93.40\% and reduces variation in the overall average, supporting the residual path's contribution beyond the main-seed result.

\begin{table}[t]
\centering
\caption{\textbf{Fairness of comparisons.} Video-level AUROC (\%). GenD and DeMamba columns change only the listed setting from the shared protocol (\cref{app:protocol}) and use a fixed seed; shared-protocol results are given in \cref{tab:crossdataset_main_table} of the main paper. The semantic path reports mean $\pm$ std over five seeds, comparable to MARS in~\cref{tab:ablation-training-seeds} of the main paper. Avg.\ gives equal weight to the seven datasets.}
\label{tab:ap-fairness}
\resizebox{\linewidth}{!}{%
\begin{tabular}{l c cc c}
\toprule
 & GenD & \multicolumn{2}{c}{DeMamba} & MARS \\
\cmidrule(lr){2-2}\cmidrule(lr){3-4}\cmidrule(lr){5-5}
Dataset & CE only, 3 ep & lr $10^{-5}$, 3 ep & lr $10^{-5}$, 10 ep & Semantic path only \\
\midrule
\multicolumn{5}{l}{\textit{Source test}} \\
FF++     & 98.00 & 53.21 & 83.04 & 95.67{\scriptsize$\pm$4.67} \\
\midrule
\multicolumn{5}{l}{\textit{Facial manipulation}} \\
CelebDF  & 90.70 & 52.90 & 70.18 & 83.83{\scriptsize$\pm$11.87} \\
DF-TIMIT & 99.40 & 49.81 & 93.19 & 99.42{\scriptsize$\pm$0.13} \\
UADFV    & 83.90 & 54.02 & 78.43 & 78.81{\scriptsize$\pm$15.28} \\
HiFiFace & 86.00 & 50.95 & 54.54 & 82.91{\scriptsize$\pm$9.34} \\
\midrule
\multicolumn{5}{l}{\textit{Generated video}} \\
GenVideo & 93.60 & 60.88 & 28.51 & 93.06{\scriptsize$\pm$2.86} \\
DF26     & 82.76 & 51.27 & 48.07 & 78.12{\scriptsize$\pm$12.03} \\
\midrule
Avg.     & 90.62 & 53.29 & 65.14 & 87.40{\scriptsize$\pm$3.93} \\
\bottomrule
\end{tabular}}
\end{table}

\begin{table}[t]
\centering\small
\setlength{\tabcolsep}{6pt}
\caption{\textbf{Learned gate versus fixed fusion weights.}
Results with a fixed seed and three epochs. Each fixed row replaces
the learned semantic-path weight $w$ with a constant. \FGdef. Best results are \textbf{bold} and second-best results are \underline{underlined} in each column.}
\label{tab:ap-gate-fixed}
\begin{tabular}{@{}lcccc@{}}
\toprule
Fusion weight $w$ & F & G & Avg. & EER\,$\downarrow$\\
\midrule
\grp{5}{Fixed}
0 (residual path) & 90.62 & 76.21 & 87.66 & 18.39\\
0.25 & 92.44 & 81.69 & 90.33 & 15.32\\
0.5 (equal average) & \textbf{92.99} & 84.25 & \underline{91.35} & \underline{13.65}\\
0.75 & 92.57 & \textbf{84.72} & 91.19 & 14.17\\
1 (semantic path) & 89.41 & 83.78 & 89.00 & 15.44\\
\midrule
\rowcolor{oursrow}
\textbf{Learned gate (MARS)} & \underline{92.95} & \underline{84.63} & \textbf{91.43} & \textbf{13.26}\\
\bottomrule
\end{tabular}
\end{table}

\begin{table}[t]
\centering\small
\setlength{\tabcolsep}{6pt}
\caption{\textbf{Residual streams and scales.}
Results with a fixed seed and three epochs. Each row changes
only the listed setting from the highlighted default
(seven streams, $\tau \in \{0.5, 1, 2, 4\}$). \FGdef. Best results are \textbf{bold} and second-best results
are \underline{underlined} in each column.}
\label{tab:ap-streams-scales}
\begin{tabular}{@{}llccc@{}}
\toprule
Residual streams & Scales $\tau$ & F & G & Avg.\\
\midrule
\grp{5}{Streams and scales}
\rowcolor{oursrow}
7 & $\{0.5, 1, 2, 4\}$ & \textbf{92.95} & \underline{84.63} & \textbf{91.43}\\
3 & $\{1, 2\}$ & \underline{90.58} & 79.57 & 88.68\\
None (appearance only) & -- & 86.23 & \textbf{90.56} & \underline{88.91}\\
\bottomrule
\end{tabular}
\end{table}

\begin{table}[t]
\centering\footnotesize
\setlength{\tabcolsep}{4pt}
\caption{\textbf{Temporal localization on spliced FF++ clips.}
Frame AUROC measures how well manipulated frames rank
above authentic frames. Top-1 is the percentage of clips
whose highest-scoring frame is manipulated, and Chance
is the manipulated fraction of the clip. All metrics
are percentages.}
\label{tab:localization}
\resizebox{\linewidth}{!}{%
\begin{tabular}{@{}lcccc@{}}
\toprule
Group & Clips & Frame AUROC & Top-1 & Chance\\
\midrule
\rowcolor{oursrow}
Overall & 3,360 & 80.73 & 93.30 & 37.50\\
\midrule
\multicolumn{5}{l}{\textit{Segment length}}\\
16 frames & 1,680 & 79.15 & 90.06 & 25.00\\
32 frames & 1,680 & 82.30 & 96.55 & 50.00\\
\midrule
\multicolumn{5}{l}{\textit{Segment position}}\\
Start & 1,120 & 79.65 & 92.68 & 37.50\\
Middle & 1,120 & 82.75 & 94.55 & 37.50\\
End & 1,120 & 79.78 & 92.68 & 37.50\\
\midrule
\multicolumn{5}{l}{\textit{Manipulation}}\\
Deepfakes & 840 & 84.67 & 96.19 & 37.50\\
Face2Face & 840 & 79.15 & 93.57 & 37.50\\
FaceSwap & 840 & 76.48 & 94.17 & 37.50\\
NeuralTextures & 840 & 82.61 & 89.29 & 37.50\\
\bottomrule
\end{tabular}%
}
\end{table}

\subsection{Semantic-Path Frame Sampling}
We vary the semantic path's input from 8 to 64 frames while
keeping the residual path at 64 frames and training for three
epochs (\cref{tab:semantic-frames}). Increasing semantic
sampling from 8 to 16 frames raises average AUROC from
90.76\% to 91.43\%, while using all 64 reaches 92.05\%.
However, average EER is lowest with 16 frames at 13.26\%,
compared with 14.33\% using 64. The default balances detection
performance with fewer frames processed through the complete
encoder.

\begin{table*}[t]
\centering\scriptsize
\setlength{\tabcolsep}{3pt}
\caption{\textbf{Semantic-path frame sampling.}
Video-level AUROC / EER (\%) with a fixed seed and three epochs.
The residual path processes 64 frames in every setting.
Avg.\ gives equal weight to seven datasets.
The default is highlighted.}
\label{tab:semantic-frames}
\resizebox{\textwidth}{!}{%
\begin{tabular}{@{}l*{8}{c}@{}}
\toprule
Semantic frames & FF++ & CelebDF & DF-TIMIT & UADFV
& HiFiFace & GenVideo & DF26 & Avg.\\
\midrule
8
& 99.01 / 4.37
& 91.05 / 16.81
& 99.87 / 1.41
& 97.38 / 12.24
& 84.49 / 20.71
& 92.53 / 14.10
& 70.96 / 34.02
& 90.76 / 14.81\\
\rowcolor{oursrow}
\textbf{16 (default)}
& 98.98 / 4.46
& 91.22 / 14.80
& 99.95 / 0.94
& 96.58 / 8.16
& 84.05 / 22.50
& 93.44 / 12.20
& 75.81 / 29.78
& 91.43 / 13.26\\
32
& 99.26 / 4.20
& 91.43 / 15.23
& 99.88 / 1.80
& 97.25 / 12.24
& 85.97 / 21.43
& 92.82 / 14.19
& 75.13 / 30.69
& 91.68 / 14.25\\
64
& 99.18 / 5.09
& 91.56 / 16.09
& 99.91 / 1.72
& 97.42 / 13.27
& 85.92 / 20.71
& 93.41 / 12.63
& 76.92 / 30.81
& 92.05 / 14.33\\
\bottomrule
\end{tabular}}
\end{table*}

\subsection{Learned Gate versus Fixed Fusion Weights}
We replace the learned gate with constant weights at inference, keeping both trained paths and their margin normalization unchanged (\cref{tab:ap-gate-fixed}). Fixed weights favor different categories: equal weighting gives the highest facial AUROC, while $w=0.75$ gives the highest generated-video AUROC. The learned gate achieves the highest overall AUROC at 91.43\% and the lowest EER at 13.26\%, compared with 91.35\% and 13.65\% for equal weighting. We retain the learned gate because it allows the contribution of each path to vary across clips without selecting a fixed mixture weight, although its overall improvement over equal weighting is negligible.

\subsection{Residual Streams and Scales}
Reducing the seven residual streams to three using $\tau\in\{1,2\}$ lowers average AUROC from 91.43\% to 88.68\% (\cref{tab:ap-streams-scales}). Removing all streams raises generated-video AUROC to 90.56\%, but lowers facial AUROC to 86.23\%. This variant still models appearance across frames, and its stronger generated-video performance echoes the semantic-only results. Together, the ablations suggest that residual comparisons favor facial detection, while appearance supports generated-video transfer; combining them yields the highest overall average.

\begin{figure*}[t]
  \centering
  \begin{subfigure}[t]{0.5\textwidth}
    \centering
    \includegraphics[width=\linewidth]{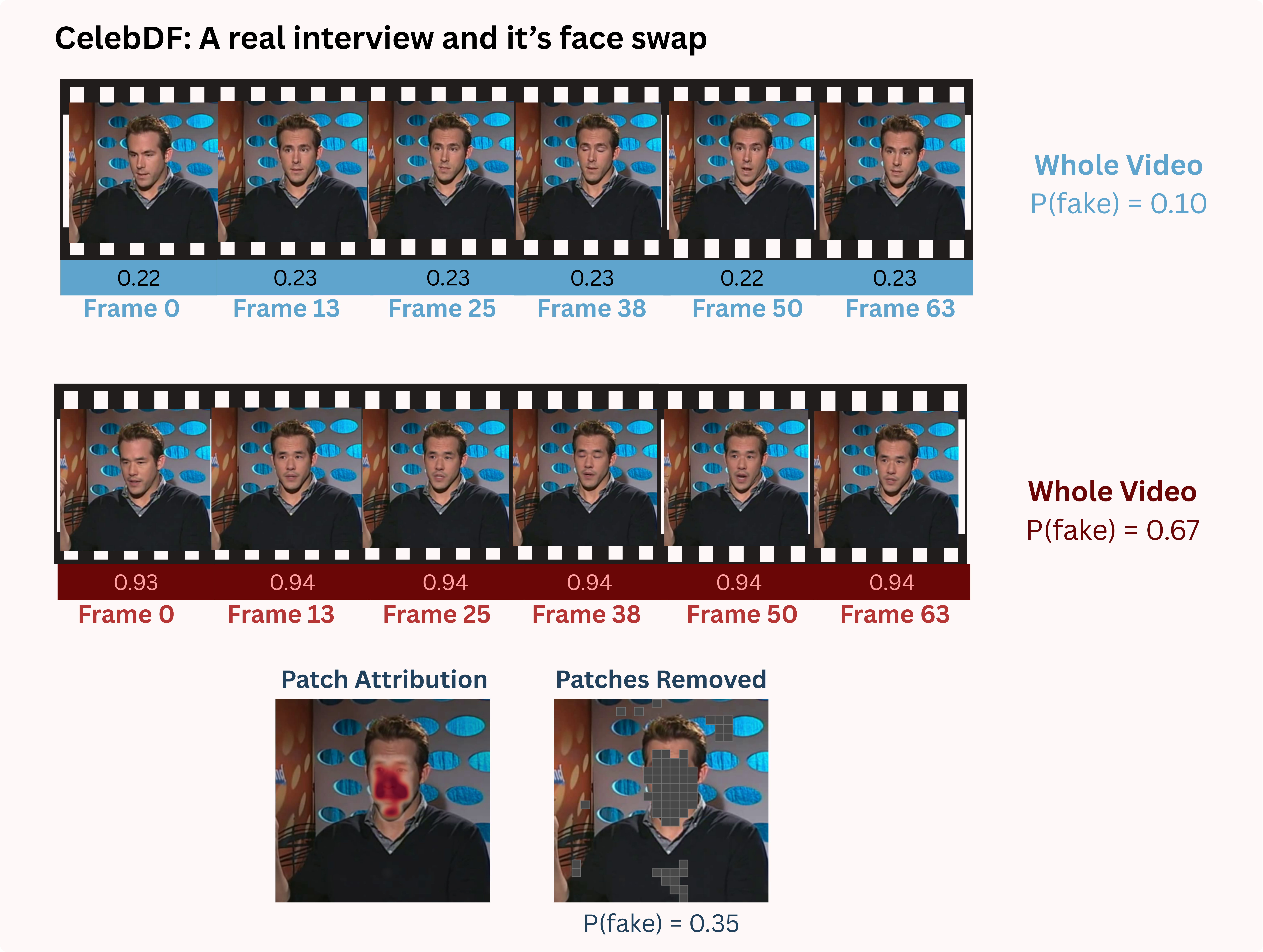}
    \caption{\textbf{Facial manipulation on Celeb-DF.}
    Authentic and face-swapped frames, clip-averaged patch attribution,
    and the prediction after removing the 64 highest-ranked patches
    per frame.}
    \label{fig:qual_celebdf}
  \end{subfigure}%
  \begin{subfigure}[t]{0.5\textwidth}
    \centering
    \includegraphics[width=\linewidth]{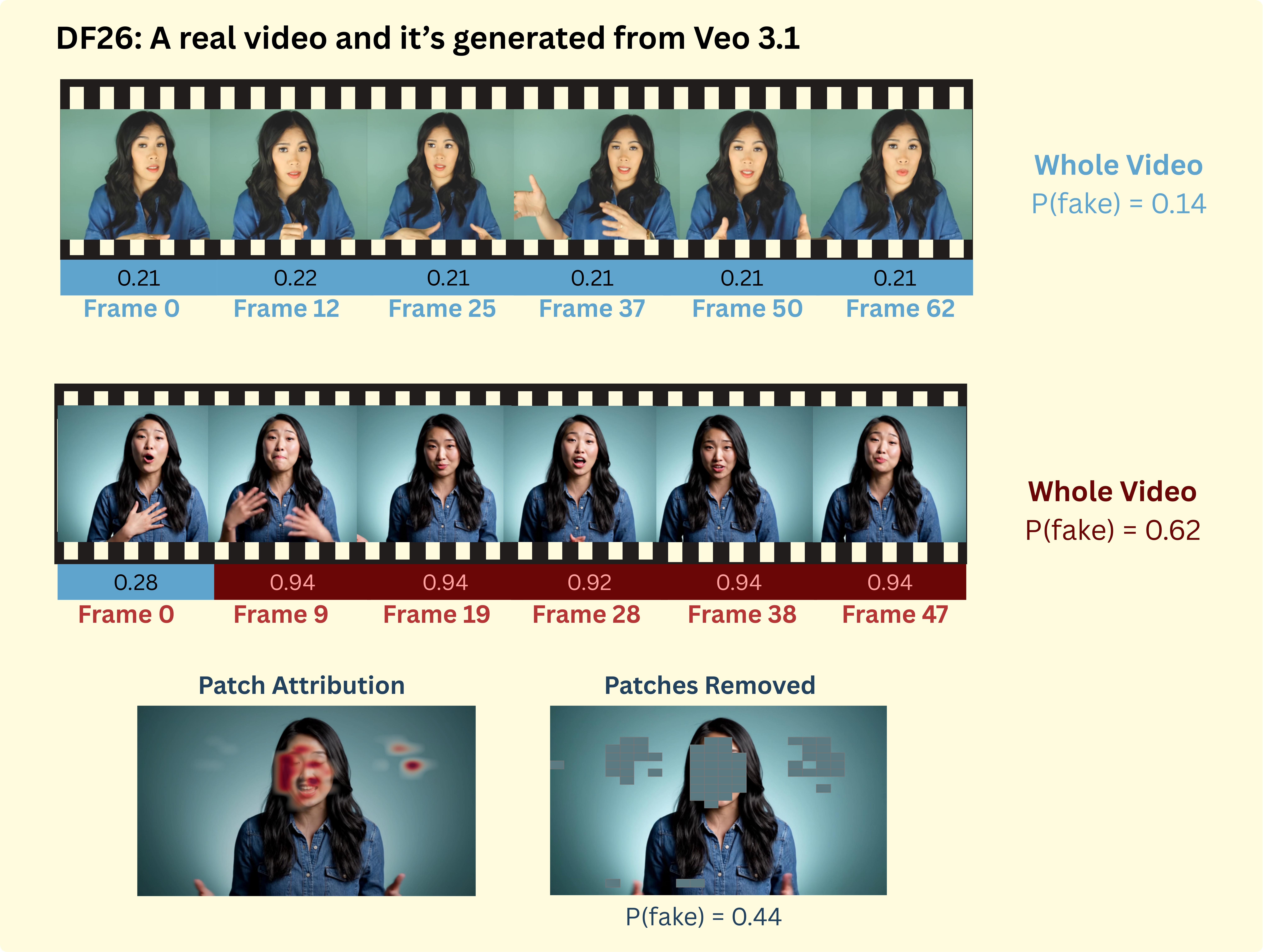}
    \caption{\textbf{Generated video on DF26.}
    Authentic and Veo~3.1 frames, clip-averaged patch attribution,
    and the prediction after removing the 64 highest-ranked patches
    per frame.}
    \label{fig:qual_df26}
  \end{subfigure}
  \caption{\textbf{Spatial evidence on facial manipulation and generated video.}}
\end{figure*}

\begin{figure*}[t]
  \centering
  \begin{subfigure}[t]{0.5\textwidth}
    \centering
    \includegraphics[width=\linewidth]{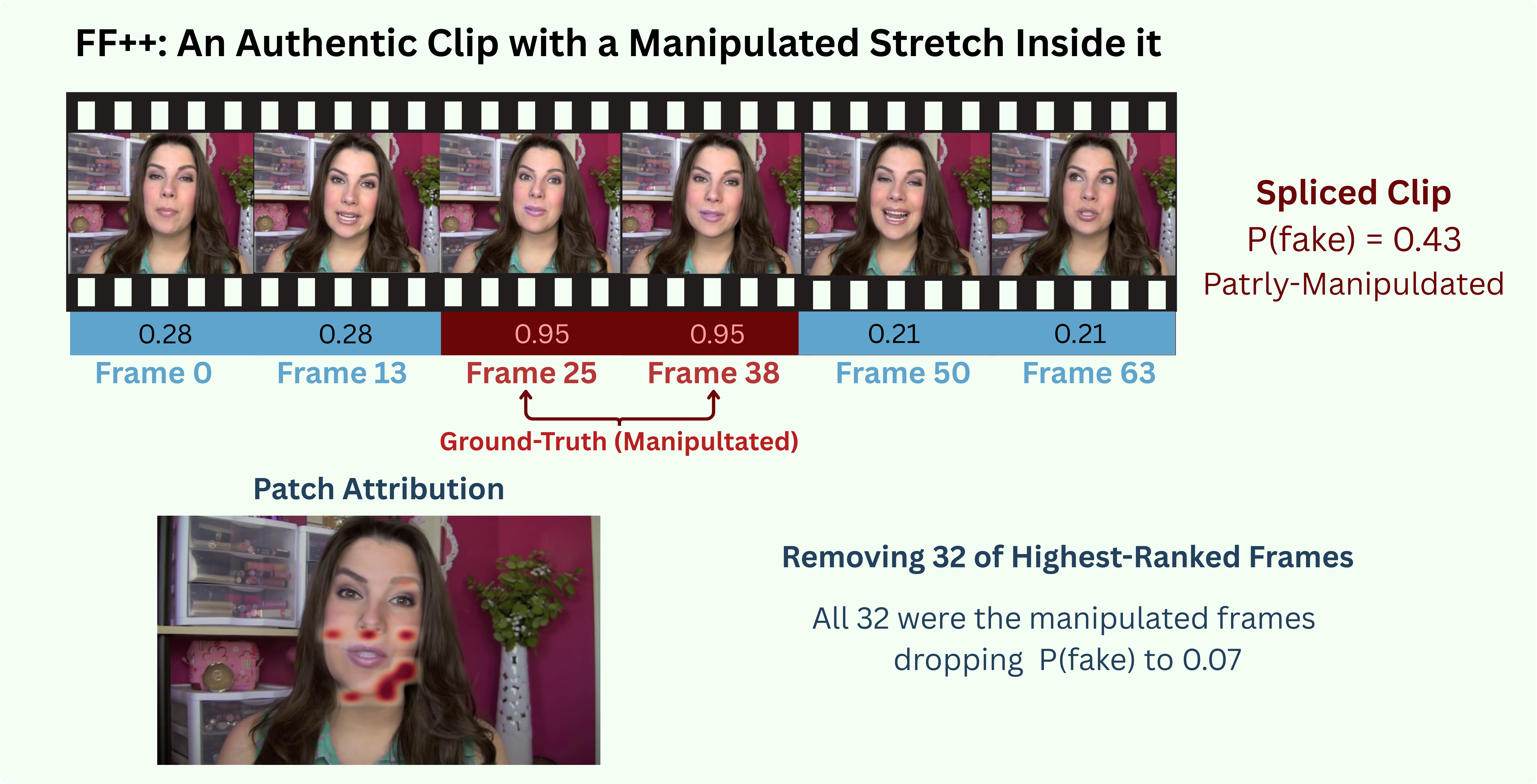}
    \caption{\textbf{Temporal localization on FF++.}
    Frames 16--47 of the illustrated clip contain a Face2Face
    manipulation and match the 32 highest-ranked frames.
    The reported overall probability is the video-level score;
    frame indices start at zero.}
    \label{fig:qual_splice}
  \end{subfigure}%
  \begin{subfigure}[t]{0.5\textwidth}
    \centering
    \includegraphics[width=\linewidth]{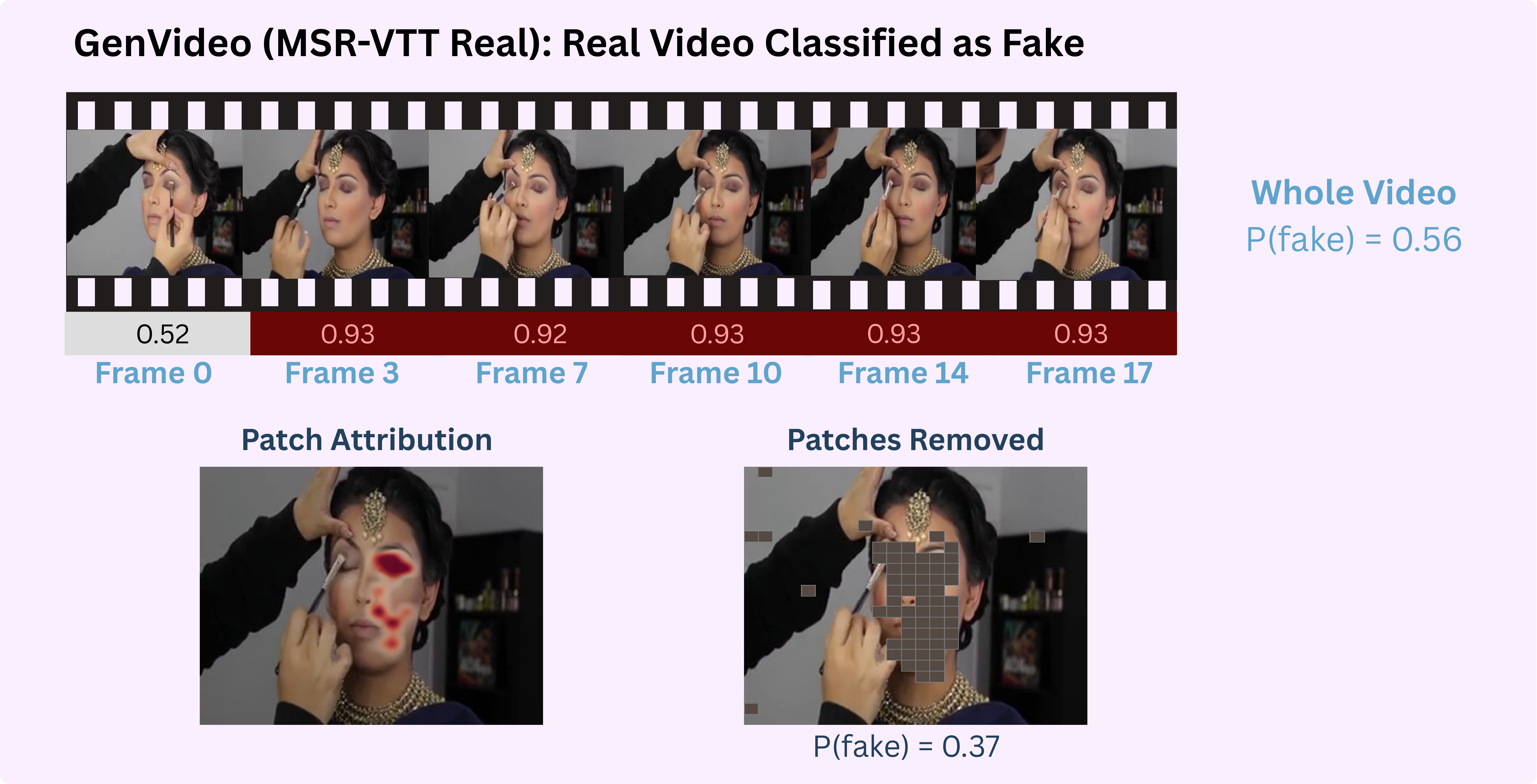}
    \caption{\textbf{False positive on an authentic GenVideo video.}
    Clip-averaged attribution highlights facial regions whose removal
    lowers the fake probability from 0.56 to 0.37.}
    \label{fig:qual_failure}
  \end{subfigure}
  \caption{\textbf{Temporal localization and a detection failure.}}
\end{figure*}
\subsection{Temporal Localization}
\label{app:temporal-localization}

\paragraph{Setup.}
We pair the 140 FF++ test originals with their four manipulation variants, giving 560 pairs. Both videos are sampled at stride two from the same starting frame. We then insert 16 or 32 manipulated frames into a 64-frame authentic clip, placing the segment at the
start, middle, or end. This produces 3,360 clips whose manipulated frames are known.

To check frame alignment, we compare the outer 15\% of sampled frames and require a mean absolute pixel difference below 10 on the 0--255 scale. For 86 pairs with a center-cropped manipulated video, we crop the authentic video to match; all 560 pairs pass the check.
Using corresponding footage avoids introducing an unrelated scene change at the segment boundary.

\paragraph{Results.}
We compute frame AUROC within each clip and then average across clips. Top-1 measures whether the highest-scoring frame falls inside the manipulated segment.~\cref{tab:localization} shows a Top-1 rate of 93.30\%, compared with 37.50\% chance,
while the lower frame AUROC indicates that ranking the entire segment remains harder.

\subsection{Qualitative Examples and Failure Cases}
\label{app:qualitative}

We examine spatial and temporal evidence on facial manipulation,
generated video, and a false positive. Frame probabilities come
from the semantic path, while video probabilities combine both
paths and are averaged over clips. Each spatial map averages
patch contributions over the illustrated clip and is overlaid
on a representative frame.

\begin{table*}[t]
\centering\scriptsize
\setlength{\tabcolsep}{4pt}
\renewcommand{\arraystretch}{0.9}
\caption{\textbf{Cross-dataset average precision.}
Video-level AP (\%) under both training settings of the main
cross-dataset comparison. Checkmarks indicate training corpora;
Avg.\ gives equal weight to the seven datasets.
Best results are \textbf{bold} and second-best results are
\underline{underlined} within each training setting, including ties.
AP depends on class balance, so comparisons should be made
between methods on the same dataset.}
\label{tab:ap-main}
\begin{tabular}{@{}ccl*{8}{c}@{}}
\toprule
\multicolumn{2}{c}{\textbf{Train}} & & \textbf{Source test} & \multicolumn{4}{c}{\textbf{Facial manipulation}} & \multicolumn{2}{c}{\textbf{Generated video}} & \textbf{Avg.}\\
\cmidrule(lr){1-2}\cmidrule(lr){4-4}\cmidrule(lr){5-8}\cmidrule(lr){9-10}\cmidrule(lr){11-11}
FF++ & GTA-V & Method & FF++ & CelebDF & DF-TIMIT & UADFV & HiFiFace & GenVideo & DF26 & \\
\midrule
& & \multicolumn{9}{l}{\textit{Frame-level}}\\
\cmark & \xmark & ForAda~\cite{cui2025forensics} & 96.46 & 92.37 & 98.48 & 96.70 & 73.77 & 52.99 & 87.63 & 85.48\\
\cmark & \xmark & Effort~\cite{yan2024effort} & 91.87 & 73.95 & 90.90 & 94.50 & 59.48 & 52.61 & 86.40 & 78.53\\
\cmark & \xmark & GenD~\cite{yermakov2026deepfake} & 89.47 & 77.67 & 99.44 & 49.48 & 44.90 & \underline{67.39} & 92.33 & 74.38\\
\cmark & \xmark & DFD-HR~\cite{sun2026dfd} & \textbf{99.95} & \textbf{97.20} & \textbf{99.98} & \underline{97.98} & \textbf{88.77} & 67.27 & \underline{94.96} & \underline{92.30}\\
\cmidrule(lr){3-11}
& & \multicolumn{9}{l}{\textit{Clip-level}}\\
\cmark & \xmark & UNITE~\cite{kundu2025towards} & 97.38 & 88.56 & 95.35 & 97.65 & 73.39 & 64.07 & 88.30 & 86.38\\
\cmark & \xmark & DeMamba~\cite{chen2026demamba} & 80.71 & 62.24 & 66.17 & 51.51 & 50.81 & 50.11 & 90.40 & 64.56\\
\cmark & \xmark & ReStraV~\cite{interno2026ai} & 81.51 & 65.07 & 46.25 & 51.72 & 53.74 & 27.84 & 92.87 & 59.86\\
\cmidrule(lr){3-11}
\rowcolor{oursrow}
\cmark & \xmark & \textbf{MARS (Ours)} & \underline{99.80} & \underline{94.62} & \underline{99.93} & \textbf{98.29} & \underline{87.77} & \textbf{92.99} & \textbf{95.58} & \textbf{95.57}\\
\midrule
& & \multicolumn{9}{l}{\textit{Frame-level}}\\
\cmark & \cmark & ForAda~\cite{cui2025forensics} & 96.38 & 91.37 & 98.34 & 97.06 & 73.64 & 58.97 & 87.74 & 86.21\\
\cmark & \cmark & Effort~\cite{yan2024effort} & 91.69 & 73.42 & 90.92 & 93.61 & 59.29 & 55.70 & 86.78 & 78.77\\
\cmark & \cmark & GenD~\cite{yermakov2026deepfake} & 90.12 & 77.89 & 99.34 & 48.17 & 51.61 & 67.63 & 92.65 & 75.34\\
\cmark & \cmark & DFD-HR~\cite{sun2026dfd} & \textbf{99.95} & \textbf{96.70} & \underline{99.82} & 96.96 & \textbf{88.91} & \underline{79.10} & \textbf{95.83} & \underline{93.90}\\
\cmidrule(lr){3-11}
& & \multicolumn{9}{l}{\textit{Clip-level}}\\
\cmark & \cmark & UNITE~\cite{kundu2025towards} & 95.85 & 84.39 & 93.07 & \textbf{98.74} & 68.78 & 66.85 & 88.65 & 85.19\\
\cmark & \cmark & DeMamba~\cite{chen2026demamba} & 80.47 & 63.48 & 66.73 & 51.64 & 51.03 & 50.39 & 91.25 & 65.00\\
\cmark & \cmark & ReStraV~\cite{interno2026ai} & 81.48 & 66.77 & 46.18 & 51.89 & 53.47 & 28.25 & 93.04 & 60.16\\
\cmidrule(lr){3-11}
\rowcolor{oursrow}
\cmark & \cmark & \textbf{MARS (Ours)} & \underline{99.74} & \underline{94.73} & \textbf{99.97} & \underline{97.29} & \underline{85.95} & \textbf{92.60} & \underline{95.76} & \textbf{95.15}\\
\bottomrule
\end{tabular}%
\end{table*}

\begin{table*}[t]
\centering\scriptsize
\setlength{\tabcolsep}{2.5pt}
\renewcommand{\arraystretch}{0.8}
\caption{\textbf{Per-generator detection on GenVideo and DF26.}
Video-level EER (\%, lower is better) for methods trained on
FF++ + GTA-V under the shared full-frame protocol with a fixed seed
and three epochs. Each generator is evaluated against all authentic
videos from the same dataset. Best results are \textbf{bold} and
second-best results are \underline{underlined} in each column.}
\label{tab:ap-pergen-detail-eer}
\resizebox{\textwidth}{!}{%
\begin{tabular}{@{}l*{20}{c}@{}}
\toprule
& \multicolumn{10}{c}{\textbf{GenVideo}}
& \multicolumn{10}{c}{\textbf{DF26}}\\
\cmidrule(lr){2-11}\cmidrule(lr){12-21}
& \multicolumn{10}{c}{}
& \multicolumn{4}{c}{\textit{Commercial}}
& \multicolumn{6}{c}{\textit{Open-source}}\\
\cmidrule(lr){12-15}\cmidrule(lr){16-21}
Method & Sora & \shortstack{Morph\\Studio}
& \shortstack{Runway\\Gen2} & HotShot & Lavie & Show-1
& \shortstack{Moon\\Valley} & Crafter
& \shortstack{Model\\Scope} & \shortstack{Wild\\Scrape}
& \shortstack{Grok\\Imagine}
& \shortstack{Kling\\3.0}
& \shortstack{Veo\\3.1}
& \shortstack{Wan\\2.6}
& \shortstack{HV1.5\\I2V}
& \shortstack{HV1.5\\T2V}
& \shortstack{Wan2.2\\I2V}
& \shortstack{Wan2.2\\T2V}
& \shortstack{LTX2.3\\I2V}
& \shortstack{LTX2.3\\T2V}\\
\midrule
\grp{21}{Frame-level}
ForAda
& 40.84 & 32.38 & 53.20 & \underline{32.39}
& 39.20 & 38.01 & 43.58 & 36.71
& 27.11 & 33.40
& 70.67 & 62.43 & 76.26 & 63.09
& 54.43 & 43.17 & 41.33 & 53.51
& 47.42 & 61.99\\
Effort
& 51.78 & 39.99 & 51.52 & 43.28
& 46.21 & 42.15 & 51.67 & 47.78
& 35.46 & 36.75
& 66.38 & 52.90 & 70.92 & 64.89
& 53.14 & 56.09 & 51.48 & 57.20
& 51.11 & 65.68\\
GenD
& 39.26 & 42.57 & 30.83 & 55.75
& 44.22 & 51.88 & 32.90 & 28.61
& 41.15 & 40.58
& \textbf{27.91} & 30.72 & 49.04 & 42.10
& 50.55 & 41.70 & 47.60 & 50.74
& 44.46 & 43.54\\
DFD-HR
& \underline{33.55} & \underline{25.70}
& \underline{22.70} & 45.43
& \underline{32.07} & 31.54
& 37.39 & \underline{14.03}
& \underline{20.26} & \underline{27.18}
& 43.75 & \underline{26.48}
& \underline{40.11} & \underline{28.42}
& \underline{39.85} & \underline{20.30}
& \textbf{26.38} & \underline{37.08}
& \textbf{16.97} & \textbf{29.52}\\
\midrule
\grp{21}{Clip-level}
UNITE
& 39.30 & 33.76 & 45.61 & 34.15
& 33.69 & \underline{30.29}
& \underline{23.99} & 35.19
& 36.70 & 31.74
& 52.95 & 53.34 & 63.92 & 59.40
& 57.20 & 43.54 & 50.00 & 60.70
& 55.54 & 60.52\\
DeMamba
& 48.25 & 57.65 & 57.20 & 42.86
& 48.82 & 44.86 & 57.82 & 52.33
& 47.07 & 47.15
& 45.74 & 44.15 & 47.87 & 48.49
& 47.79 & 66.05 & 49.08 & 42.80
& 47.97 & \underline{32.66}\\
ReStraV
& 92.97 & 71.34 & 97.58 & 90.78
& 79.96 & 81.94 & 97.46 & 80.42
& 79.16 & 79.05
& 38.60 & 40.43 & 49.29 & 43.16
& 49.82 & 49.82 & 38.01 & 37.64
& 47.23 & 37.64\\
\midrule
\rowcolor{oursrow}
\textbf{MARS (Ours)}
& \textbf{26.40} & \textbf{13.43}
& \textbf{10.79} & \textbf{21.44}
& \textbf{12.91} & \textbf{19.71}
& \textbf{8.16} & \textbf{5.13}
& \textbf{10.02} & \textbf{13.10}
& \underline{32.75} & \textbf{24.55}
& \textbf{27.15} & \textbf{22.68}
& \textbf{30.07} & \textbf{19.56}
& \underline{33.39} & \textbf{32.84}
& \underline{33.03} & 36.72\\
\bottomrule
\end{tabular}}
\end{table*}

For Celeb-DF and DF26
(\cref{fig:qual_celebdf,fig:qual_df26}),
removing the 64 highest-ranked patches per frame lowers the fake
probability from 0.67 to 0.35 and from 0.62 to 0.44, respectively,
compared with 0.65 and 0.62 after random removal. These examples
show that the highlighted regions contribute to the prediction.

In the partially manipulated FF++ video
(\cref{fig:qual_splice}), the 32 highest-ranked frames in
the illustrated clip match the inserted Face2Face segment.
Removing them lowers the video-level fake probability from
0.43 to 0.07, compared with 0.42 for random removal.
This indicates that the frame ranking identifies the manipulated
interval and that this interval contributes to the fake score.

The authentic makeup tutorial (\cref{fig:qual_failure}) shows
that MARS can mistake genuine facial appearance for manipulation.
Removing the 64 highest-ranked patches per frame lowers the
false-positive score from 0.56 to 0.37, while random removal leaves
it at 0.56 on average over five draws. This suggests that the
highlighted facial regions contribute to the error, showing how
the evidence can help inspect an incorrect prediction even when
the detector fails to distinguish authentic appearance from
forgery traces.

\section{Detailed Experimental Results}
\label{app:results}

\subsection{Average Precision}
\cref{tab:ap-main} reports AP, which measures how many of the videos ranked as fake are actually fake. AP follows the same pattern as AUROC: MARS achieves the highest average in both training settings (95.57 and 95.15), followed by DFD-HR (92.30 and 93.90), while the remaining methods trail by at least 5.9 points. The largest gap appears on GenVideo, where MARS reaches 92.99 AP against at most 79.10 for the baselines. 
In contrast, DF26 compresses the differences between methods: since 90\% of its videos are fake, a random detector already reaches an AP of about 90, and even DeMamba, near chance in AUROC, scores above 90. Overall, AP confirms the AUROC ranking, but on imbalanced sets such as DF26 it largely reflects the share of fake videos, which makes AUROC and EER the more informative metrics there.

\begin{table*}[t]
\centering\scriptsize
\setlength{\tabcolsep}{3pt}
\caption{\textbf{Component ablation across datasets.}
Video-level AUROC and EER (\%) under matched training
with a fixed seed and three epochs. Avg.\ gives equal weight
to the seven datasets. Params denotes trainable parameters
in millions.  Best results are \textbf{bold} and
second-best results are \underline{underlined} in each column.}
\label{tab:ap-ablation-training}
\resizebox{\textwidth}{!}{%
\begin{tabular}{@{}lc*{16}{c}@{}}
\toprule
& & \multicolumn{2}{c}{\textbf{Source test}} & \multicolumn{8}{c}{\textbf{Facial manipulation}} & \multicolumn{4}{c}{\textbf{Generated video}} & \multicolumn{2}{c}{\textbf{Avg.}}\\
\cmidrule(lr){3-4}\cmidrule(lr){5-12}\cmidrule(lr){13-16}\cmidrule(lr){17-18}
&  & \multicolumn{2}{c}{FF++} & \multicolumn{2}{c}{CelebDF} & \multicolumn{2}{c}{DF-TIMIT} & \multicolumn{2}{c}{UADFV} & \multicolumn{2}{c}{HiFiFace} & \multicolumn{2}{c}{GenVideo} & \multicolumn{2}{c}{DF26} & & \\
\cmidrule(lr){3-4}\cmidrule(lr){5-6}\cmidrule(lr){7-8}\cmidrule(lr){9-10}\cmidrule(lr){11-12}\cmidrule(lr){13-14}\cmidrule(lr){15-16}
Variant & Params & AUROC$\uparrow$ & EER$\downarrow$ & AUROC$\uparrow$ & EER$\downarrow$ & AUROC$\uparrow$ & EER$\downarrow$ & AUROC$\uparrow$ & EER$\downarrow$ & AUROC$\uparrow$ & EER$\downarrow$ & AUROC$\uparrow$ & EER$\downarrow$ & AUROC$\uparrow$ & EER$\downarrow$ & AUROC$\uparrow$ & EER$\downarrow$ \\
\midrule
\rowcolor{oursrow}
\textbf{Full MARS} & 67.54 & 98.98 & 4.46 & \underline{91.22} & \textbf{14.80} & 99.95 & \underline{0.94} & 96.58 & 8.16 & 84.05 & 22.50 & 93.44 & \underline{12.20} & 75.81 & \underline{29.78} & \textbf{91.43} & \textbf{13.26} \\
\midrule
\grp{18}{Paths}
Residual path only & 67.43 & 98.07 & 7.05 & 82.42 & 25.29 & 99.26 & 4.22 & 97.79 & \textbf{6.12} & 77.04 & 27.86 & 74.76 & 31.45 & 64.23 & 40.13 & 84.80 & 20.30 \\
Semantic path only & 0.11 & 97.83 & 7.32 & 89.16 & 18.67 & 99.27 & 2.66 & 83.80 & 12.24 & \textbf{87.34} & \underline{20.00} & \underline{93.59} & 14.25 & \textbf{80.68} & \textbf{26.59} & 90.24 & 14.53 \\
\midrule
\grp{18}{Residual streams}
w/o appearance & 66.36 & \underline{99.09} & 4.82 & 90.51 & 17.38 & 99.67 & 1.95 & \underline{98.21} & 8.16 & \underline{86.38} & \textbf{19.29} & \textbf{93.63} & \textbf{11.81} & 71.49 & 33.59 & \underline{91.28} & \underline{13.86} \\
w/o spatial streams & 66.65 & \textbf{99.56} & \textbf{2.68} & 90.11 & 17.68 & \underline{99.98} & 1.02 & 93.29 & 14.29 & 58.80 & 43.57 & 71.39 & 35.29 & 74.61 & 31.76 & 83.96 & 20.90 \\
w/o temporal streams & 66.65 & 98.62 & 5.54 & 87.10 & 19.68 & 99.91 & 1.33 & 97.00 & \underline{7.14} & 71.08 & 33.57 & 71.75 & 34.50 & \underline{77.03} & 30.91 & 86.07 & 18.95 \\
w/o first difference & 67.24 & 99.07 & \underline{3.39} & \textbf{91.89} & \underline{15.08} & 99.90 & 1.80 & 97.38 & 10.20 & 82.28 & 23.57 & 90.03 & 15.99 & 73.12 & 32.06 & 90.52 & 14.59 \\
\midrule
\grp{18}{Semantic norms}
Frozen norms (encoder fully frozen) & 67.44 & 98.53 & 6.16 & 87.49 & 18.83 & \textbf{99.99} & \textbf{0.23} & \textbf{99.08} & \textbf{6.12} & 72.90 & 31.07 & 40.30 & 58.67 & 64.44 & 39.37 & 80.39 & 22.92 \\
\bottomrule
\end{tabular}
}
\end{table*}
\subsection{Generator-Level EER}
As shown in~\cref{tab:ap-pergen-detail-eer}, MARS achieves the lowest EER on all ten GenVideo generators and on six of the ten DF26 groups, while DFD-HR remains stronger on Wan2.2 I2V and both LTX2.3 modes, and GenD on Grok Imagine. DF26 is harder than GenVideo for all methods, with the lowest EER above 20 on eight of its ten groups, compared with two of ten on GenVideo. Interestingly, the hardest generators differ between detectors: MARS has a higher mean EER on the open-source DF26 groups than on the commercial ones (30.94 vs.\ 26.78), whereas DFD-HR shows the opposite trend (28.35 vs.\ 34.69), suggesting that the two methods rely on partly different cues. 
Overall, we observe that the advantage of MARS holds across the GenVideo generators but not across the more recent DF26 generators, where no single detector is best on every group.

\subsection{Component Ablation by Dataset}
The per-dataset results reveal where each component contributes (\cref{tab:ap-ablation-training}). The semantic path alone performs similarly to full MARS on GenVideo and better on DF26, but lowers UADFV AUROC from 96.58\% to 83.80\%.
Removing spatial DoG reduces HiFiFace AUROC from 84.05\%
to 58.80\% and GenVideo from 93.44\% to 71.39\%;
removing temporal DoG also lowers GenVideo to 71.75\%.
Freezing the semantic LayerNorms has its largest effect on GenVideo, reducing AUROC to 40.30\%.
These differences show that component benefits vary across datasets, while full MARS achieves the highest seven-dataset average among the evaluated variants.

\subsection{Training Duration and Clip Length by Dataset}
\cref{tab:duration} provides the per-dataset MARS results
for training duration and clip length. Shorter clips produce
more training batches per epoch and more clip predictions per
video, so the clip-length comparison changes both the temporal
window and the number of optimizer steps.

\begin{table*}[t]
\centering\scriptsize
\setlength{\tabcolsep}{3pt}
\caption{\textbf{Training duration and clip length for MARS.}
Video-level AUROC / EER (\%) with a fixed seed.
The default uses three epochs, 64-frame clips, and
16 semantic frames. Steps counts optimizer updates over
the full run; Avg.\ gives equal weight to seven datasets.}
\label{tab:duration}
\resizebox{\textwidth}{!}{%
\begin{tabular}{@{}lr*{8}{c}@{}}
\toprule
Setting & Steps & FF++ & CelebDF & DF-TIMIT & UADFV
& HiFiFace & GenVideo & DF26 & Avg.\\
\midrule
\rowcolor{oursrow}
\textbf{MARS (Ours): default}
& 1,449
& 98.98 / 4.46
& 91.22 / 14.80
& 99.95 / 0.94
& 96.58 / 8.16
& 84.05 / 22.50
& 93.44 / 12.20
& 75.81 / 29.78
& 91.43 / 13.26\\
\midrule
\multicolumn{10}{l}{\textit{Training duration (64-frame clips)}}\\
5 epochs
& 2,415
& 99.62 / 2.86
& 92.29 / 14.08
& 99.90 / 1.33
& 96.58 / 8.16
& 86.88 / 18.57
& 92.26 / 13.08
& 74.38 / 31.32
& 91.70 / 12.77\\
10 epochs
& 4,830
& 99.38 / 4.29
& 90.81 / 15.81
& 99.91 / 1.41
& 96.00 / 8.16
& 84.71 / 21.79
& 95.31 / 9.71
& 74.75 / 31.10
& 91.55 / 13.18\\
\midrule
\multicolumn{10}{l}{\textit{Clip length (three epochs)}}\\
8 frames
& 10,410
& 99.40 / 3.57
& 92.55 / 14.51
& 99.91 / 1.56
& 95.84 / 10.20
& 86.30 / 21.43
& 94.44 / 11.80
& 79.11 / 27.25
& 92.50 / 12.90\\
16 frames
& 5,292
& 99.32 / 4.29
& 92.73 / 13.51
& 99.96 / 0.94
& 95.63 / 11.22
& 86.71 / 21.43
& 91.22 / 16.92
& 78.26 / 28.15
& 91.98 / 13.78\\
32 frames
& 2,727
& 99.36 / 4.29
& 93.19 / 13.51
& 99.88 / 0.94
& 97.33 / 13.27
& 85.90 / 21.07
& 91.66 / 16.26
& 76.50 / 30.21
& 91.97 / 14.22\\
\bottomrule
\end{tabular}}
\end{table*}

\begin{table*}[t]
\centering
\setlength{\tabcolsep}{2.2pt}
\caption{\textbf{Robustness across datasets.}
Video-level AUROC, EER, and AP (\%), averaged over the 18
JPEG, blur, downscaling, and noise settings in
~\cref{fig:robustness}. Mean gives equal weight to the seven
datasets. Best results are \textbf{bold} and second-best results
are \underline{underlined} within each metric, including ties.}
\label{tab:robust-summary}
\resizebox{\textwidth}{!}{%
\begin{tabular}{@{}l|*{7}{ccc|}ccc@{}}
\toprule
\multirow{2}{*}{Method}
& \multicolumn{3}{c|}{FF++} & \multicolumn{3}{c|}{CelebDF} & \multicolumn{3}{c|}{DF-TIMIT} & \multicolumn{3}{c|}{UADFV}
& \multicolumn{3}{c|}{HiFiFace} & \multicolumn{3}{c|}{GenVideo} & \multicolumn{3}{c|}{DF26} & \multicolumn{3}{c}{Mean}\\
\cmidrule(lr){2-4}\cmidrule(lr){5-7}\cmidrule(lr){8-10}\cmidrule(lr){11-13}
\cmidrule(lr){14-16}\cmidrule(lr){17-19}\cmidrule(lr){20-22}\cmidrule(l){23-25}
& AUC$\uparrow$ & EER$\downarrow$ & AP$\uparrow$ & AUC$\uparrow$ & EER$\downarrow$ & AP$\uparrow$
& AUC$\uparrow$ & EER$\downarrow$ & AP$\uparrow$ & AUC$\uparrow$ & EER$\downarrow$ & AP$\uparrow$
& AUC$\uparrow$ & EER$\downarrow$ & AP$\uparrow$ & AUC$\uparrow$ & EER$\downarrow$ & AP$\uparrow$
& AUC$\uparrow$ & EER$\downarrow$ & AP$\uparrow$ & AUC$\uparrow$ & EER$\downarrow$ & AP$\uparrow$\\
\midrule
\grp{25}{Frame-level}
ForAda & 82.35 & 25.94 & 95.06 & \underline{78.28} & \underline{28.08} & \underline{86.64} & \textbf{89.64} & \textbf{17.68} & \textbf{94.29} & \textbf{92.03} & \textbf{13.04} & \textbf{93.65} & \textbf{70.30} & \underline{35.95} & \underline{69.61} & 64.31 & 39.19 & 60.03 & 38.45 & 57.20 & 87.09 & 73.62 & 31.01 & \underline{83.77} \\
Effort & 69.75 & 36.25 & 90.84 & 56.77 & 44.57 & 71.78 & 73.33 & 32.93 & 84.87 & 83.93 & 23.13 & 84.54 & 58.00 & 44.58 & 58.75 & 58.30 & 43.94 & 58.01 & 37.29 & 58.65 & 86.49 & 62.48 & 40.58 & 76.47 \\
GenD & 61.77 & 41.26 & 85.96 & 59.18 & 43.62 & 72.71 & 74.13 & 31.45 & 84.56 & 63.54 & 40.08 & 62.06 & 49.90 & 48.77 & 51.48 & \underline{68.54} & 36.31 & \underline{70.26} & 46.96 & 52.64 & 89.10 & 60.57 & 42.02 & 73.74 \\
DFD-HR & \textbf{90.60} & \textbf{16.18} & \textbf{97.37} & \textbf{80.57} & \textbf{27.06} & \textbf{88.27} & \underline{87.82} & \underline{18.30} & \underline{93.13} & 86.58 & 17.63 & 86.00 & 64.60 & 38.77 & 66.44 & 62.36 & 41.04 & 64.47 & 48.74 & 50.61 & 89.68 & \underline{74.47} & \underline{29.94} & 83.62 \\
\midrule
\grp{25}{Clip-level}
UNITE & 80.71 & 27.05 & 94.65 & 65.24 & 38.82 & 78.62 & 79.07 & 28.30 & 88.43 & \underline{90.21} & \underline{16.95} & \underline{90.50} & 64.37 & 39.76 & 65.57 & 66.80 & \underline{34.78} & 65.42 & 42.56 & 56.01 & 88.46 & 69.85 & 34.52 & 81.67 \\
DeMamba & 51.45 & 49.28 & 80.53 & 47.87 & 51.72 & 63.49 & 49.40 & 50.84 & 66.69 & 50.81 & 50.45 & 51.44 & 50.50 & 49.78 & 50.87 & 49.64 & 50.44 & 50.18 & 52.63 & \underline{46.26} & 91.22 & 50.33 & 49.82 & 64.92 \\
ReStraV & 52.75 & 48.54 & 81.59 & 51.94 & 48.25 & 67.54 & 19.10 & 74.91 & 51.85 & 52.06 & 48.19 & 52.99 & 51.93 & 49.31 & 52.20 & 9.39 & 83.36 & 28.21 & \underline{55.02} & 46.50 & \underline{91.74} & 41.74 & 57.01 & 60.87 \\
\midrule
\rowcolor{oursrow}
\textbf{MARS (Ours)} & \underline{87.31} & \underline{20.28} & \underline{96.27} & 75.66 & 30.21 & 84.23 & 85.33 & 20.69 & 92.01 & 87.70 & 18.31 & 87.35 & \underline{68.94} & \textbf{35.02} & \textbf{69.88} & \textbf{71.15} & \textbf{33.38} & \textbf{70.28} & \textbf{63.12} & \textbf{40.53} & \textbf{93.06} & \textbf{77.03} & \textbf{28.35} & \textbf{84.72} \\
\bottomrule
\end{tabular}%
}
\end{table*}

\subsection{Robustness by Dataset}
\label{app:robust}

\paragraph{Perturbation Settings.}
We apply each degradation to decoded frames at their native resolution, before encoder-specific resizing and normalization. All methods use their main-table checkpoints (a fixed seed), without additional training. JPEG compression uses Pillow quality levels of 80, 60, 40, 20, and 10. Following GenD~\cite{yermakov2026deepfake}, Gaussian blur uses standard deviations of 0.5, 1, 1.5, 2, and 3 pixels, with respective kernel sizes of 5, 7, 11, 13, and 19 and reflective padding.

Downscaling reduces the short side to 224, 112, or 64 pixels using bilinear interpolation with antialiasing, preserving the aspect ratio. Gaussian noise uses standard deviations of 0.01, 0.03, 0.05, 0.1, and 0.2 on pixel values in $[0,1]$, followed by clipping and 8-bit quantization. Noise is seeded by video identifier and clip index so that every method
receives the same perturbation.

\begin{figure*}[t]
\centering
\includegraphics[width=\textwidth]{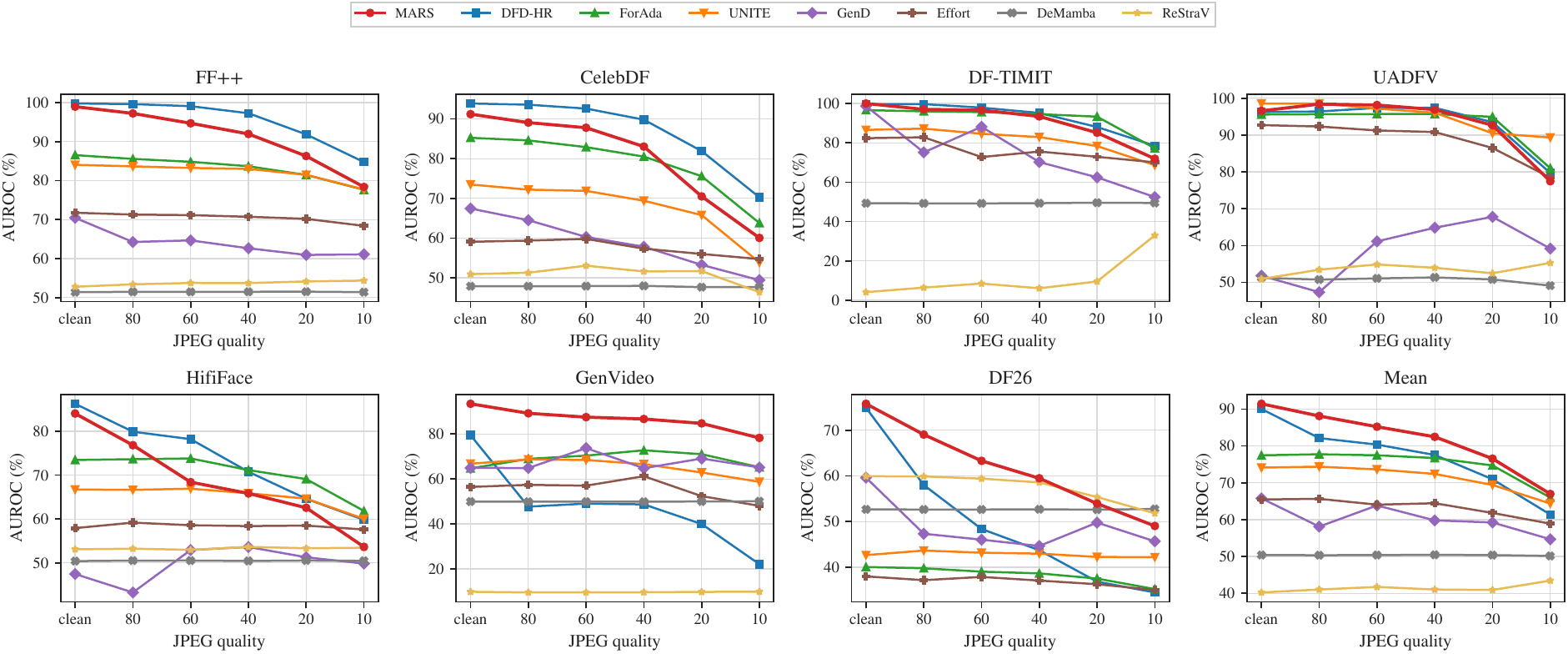}
\caption{Robustness to JPEG compression at quality 80, 60, 40, 20 and 10: video AUROC (\%) of the main-table checkpoints on each evaluation dataset and their mean. The leftmost point is the unperturbed test set.}
\label{fig:robust-auroc-jpeg}
\end{figure*}

\begin{figure*}[t]
\centering
\includegraphics[width=\textwidth]{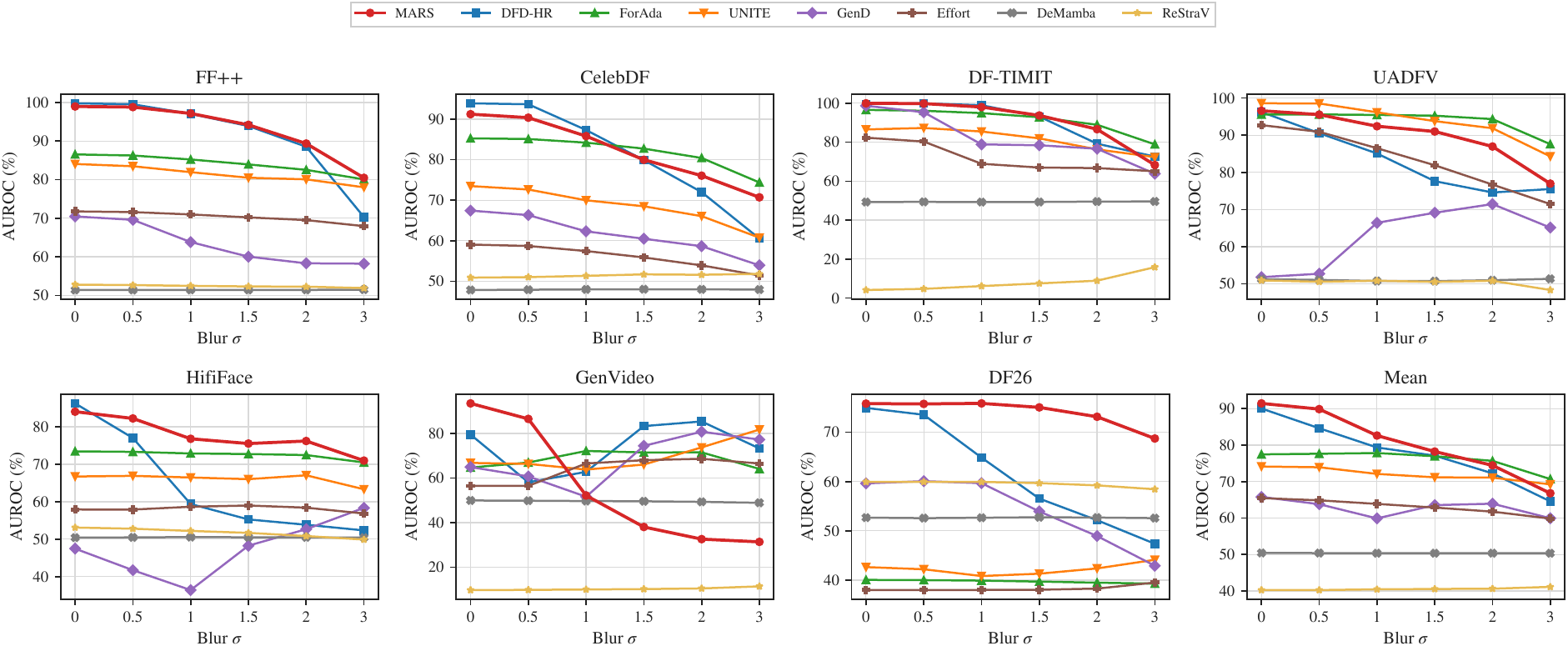}
\caption{Robustness to Gaussian blur with standard deviation 0.5, 1, 1.5, 2 and 3 pixels: video AUROC (\%) of the main-table checkpoints on each evaluation dataset and their mean. The leftmost point is the unperturbed test set.}
\label{fig:robust-auroc-blur}
\end{figure*}

\begin{figure*}[t]
\centering
\includegraphics[width=\textwidth]{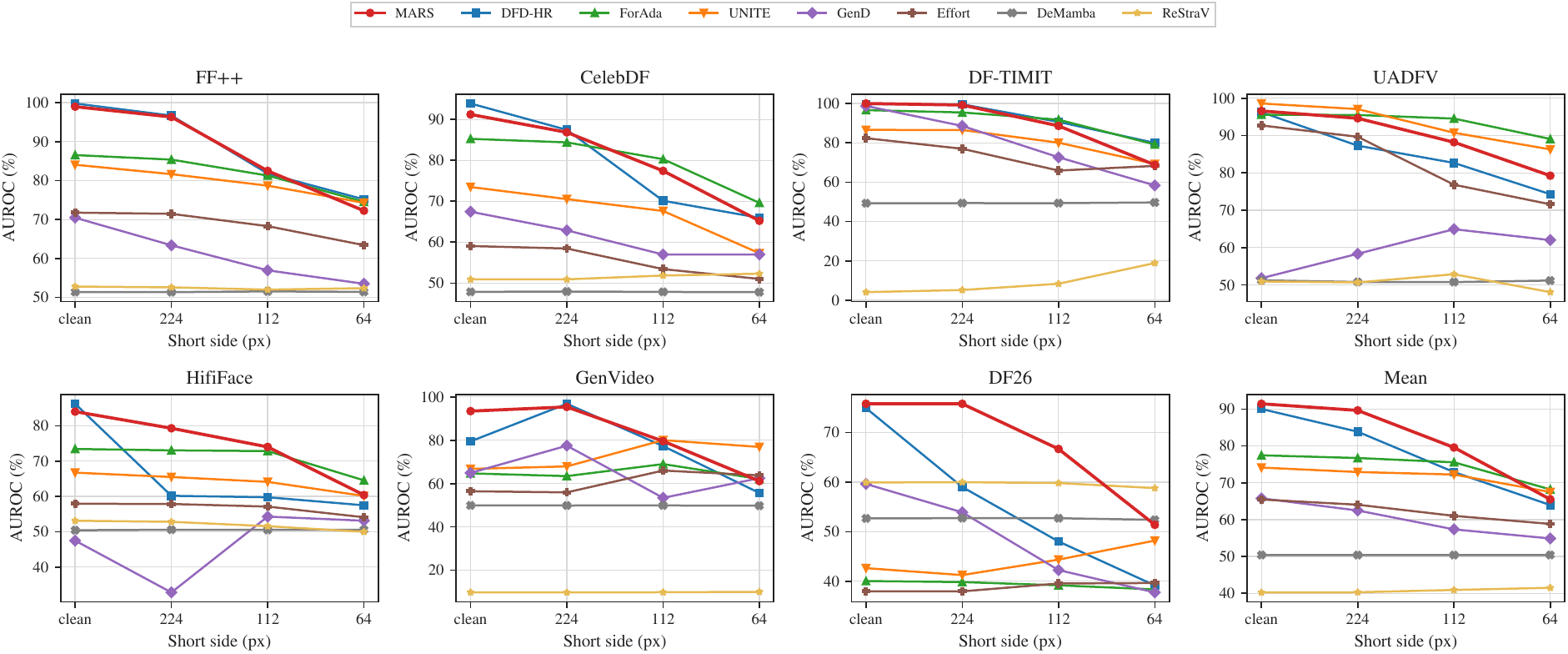}
\caption{Robustness to downscaling of the short side to 224, 112 and 64 pixels: video AUROC (\%) of the main-table checkpoints on each evaluation dataset and their mean. The leftmost point is the unperturbed test set.}
\label{fig:robust-auroc-resize}
\end{figure*}

\begin{figure*}[t]
\centering
\includegraphics[width=\textwidth]{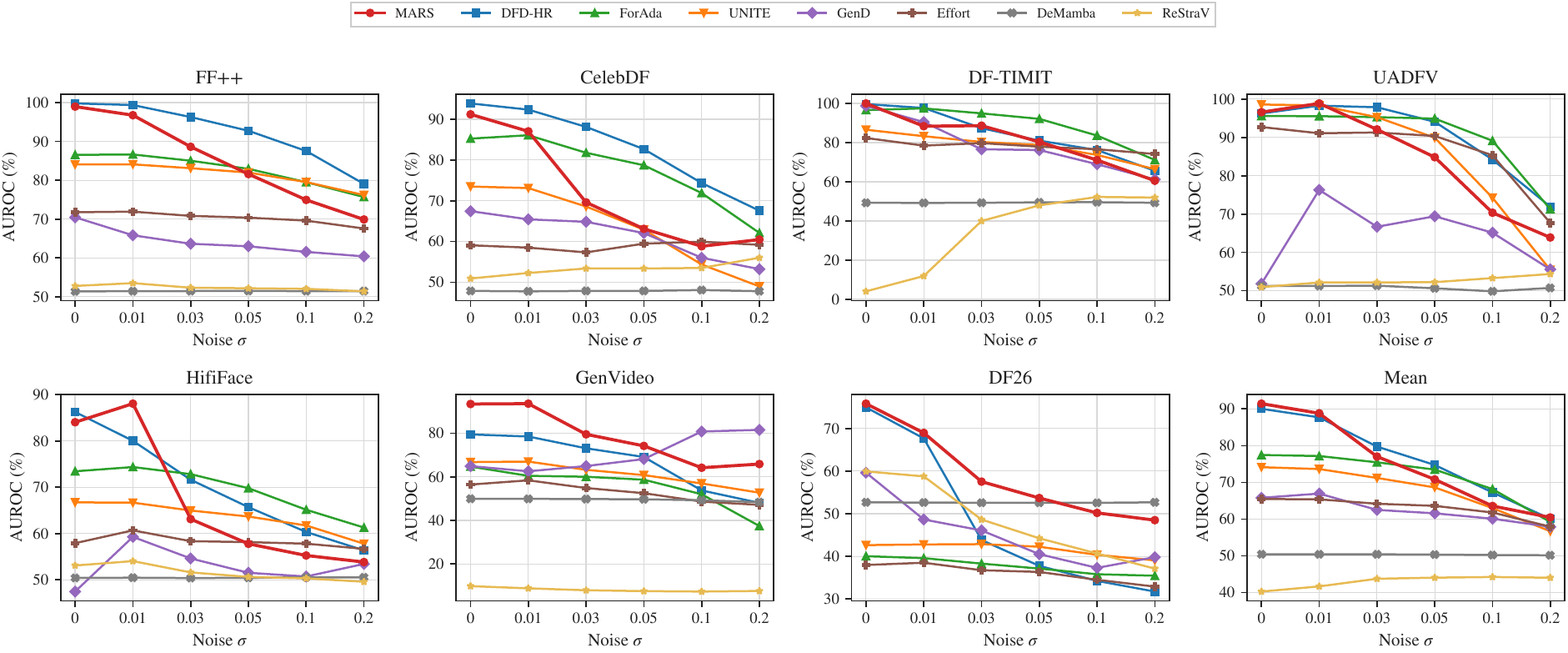}
\caption{Robustness to additive Gaussian noise with standard deviation 0.01, 0.03, 0.05, 0.1 and 0.2 on pixel values in $[0,1]$: video AUROC (\%) of the main-table checkpoints on each evaluation dataset and their mean. The leftmost point is the unperturbed test set.}
\label{fig:robust-auroc-noise}
\end{figure*}

\paragraph{Per-Dataset Results.}
~\cref{tab:robust-summary} reports metrics averaged over the 18 perturbation settings, while
~\cref{fig:robust-auroc-jpeg,fig:robust-auroc-resize,fig:robust-auroc-noise,fig:robust-auroc-blur} show AUROC at each strength for each dataset.

\subsection{Faithfulness by Dataset}

\paragraph{Patch Removal}

For each dataset, we select up to 100 manipulated clips that MARS predicts as fake and up to 100 authentic clips. We use the first clip of each selected video and require more than 16 valid frames. Selecting detected manipulated clips explains why the initial AUROC is close to 100\%.

We replace the highest-ranked $k\in\{8,16,32,64\}$ patches with each frame's mean color. Each patch covers $14\times14$ pixels, and the same positions are replaced in every frame. The random control replaces the same number of patches on the same clips using a fixed random seed. We rerun the detector after each edit and compute AUROC using one clip per selected video.

\begin{figure*}[t]
\centering
\includegraphics[width=\textwidth]{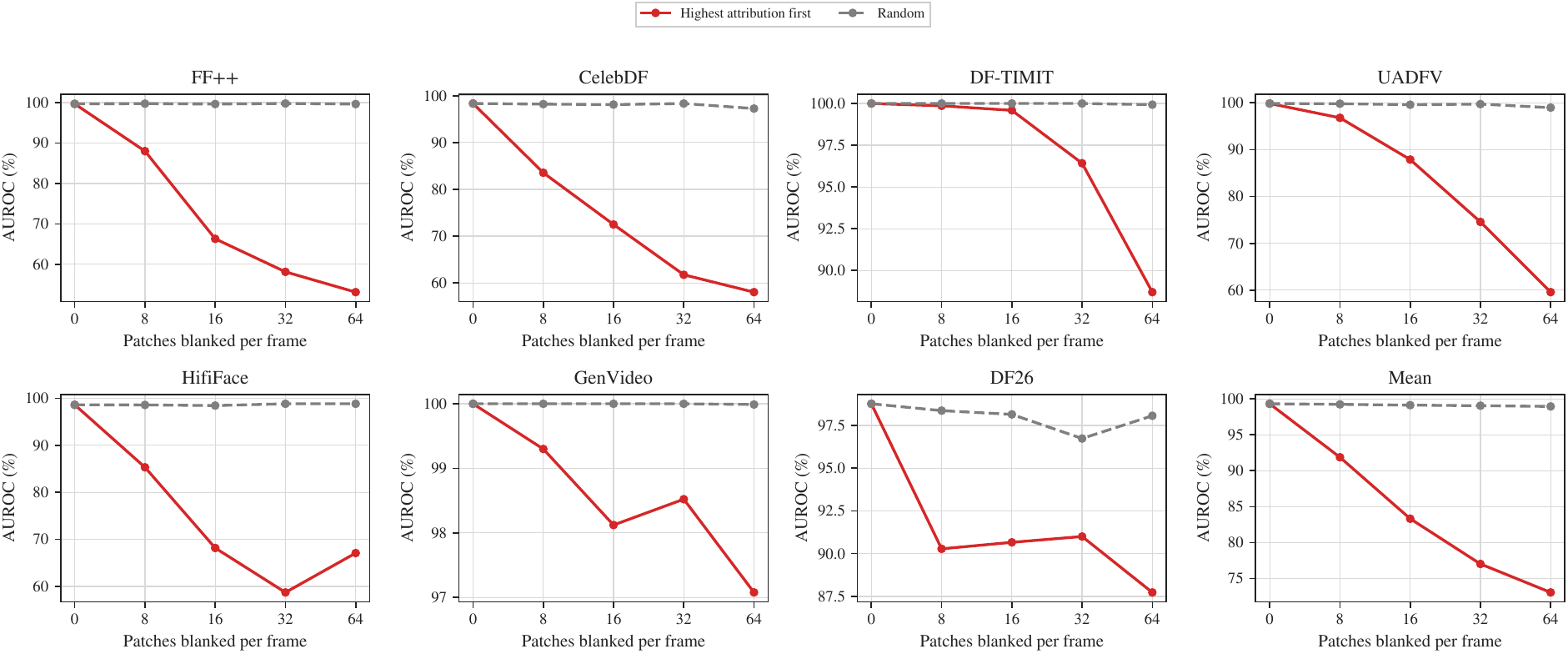}
\caption{\textbf{Patch removal by dataset.}
AUROC (\%) after replacing the highest-ranked or
random patches with each frame's mean color.
The same positions are replaced across frames,
using one clip per selected video.}
\label{fig:faith-patches}
\end{figure*}

As shown in~\cref{fig:faith-patches}, ranked removal reduces detection performance more than random removal, particularly on FF++, CelebDF, and UADFV. The effect is smaller on GenVideo, which may reflect evidence spread across more of the scene. This test measures the relevance of selected patches to the prediction, not whether they match the actual manipulated regions.

\paragraph{Frame Removal}

\begin{figure}[t]
\centering
\includegraphics[width=\linewidth]{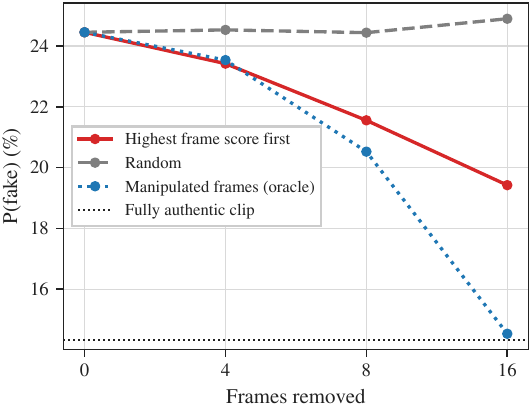}
\caption{\textbf{Frame removal on spliced FF++ clips.}
Mean fake probability (\%) after removing the
highest-scoring, random, or known manipulated frames
from 560 clips with a 16-frame manipulated segment.
The dotted line marks the fully authentic reference.}
\label{fig:faith-frames}
\end{figure}

We use clips with a manipulated segment in the middle, giving 560 clips for each segment length. For a segment of length $L$, we remove $L/4$, $L/2$, or $L$ frames using three choices: the highest frame scores, random frames, or known manipulated frames (oracle).
The remaining frames keep their original order and are packed together with an updated validity mask before MARS is run again.
For the 16-frame segments, removing the 16 highest-scoring frames lowers the mean fake probability from 24.45\% to 19.42\%, compared with 24.89\% for random removal and 14.53\% for oracle removal (\cref{fig:faith-frames}). Removing frames also changes the temporal spacing between the remaining frames and the frames sampled by the semantic path, which random removal of the same number of frames also affects, though the resulting tmeporal gaps and selected frames may differ. Ranked removal moves the prediction closer to the authentic reference, but the gap to oracle removal shows that the ranking does not identify every manipulated frame.

\subsection{Compute and Environmental Footprint}
We track the GPU-hours, energy, carbon emissions, and water use of each run with the ai-impact-accounting package~\cite{raza2026position,dia2026pypi}. \cref{tab:footprint} reports these quantities for the main runs, separately for training and for testing on the seven evaluation datasets.

\begin{table}[t]
\centering\footnotesize
\setlength{\tabcolsep}{2pt}
\caption{\textbf{Compute and environmental footprint.}
Main runs use FF++ + GTA-V, a fixed seed, and three epochs on four
NVIDIA A40 GPUs, tracked with the ai-impact-accounting package.
Energy use is measured; carbon emissions are estimated from
energy use and carbon intensity. Water use is estimated using
the default water-use-effectiveness range and reported as
low--high. Testing totals cover the seven evaluation datasets.}
\label{tab:footprint}
\resizebox{\linewidth}{!}{%
\begin{tabular}{@{}l cccc ccc@{}}
\toprule
& \multicolumn{4}{c}{Training} & \multicolumn{3}{c}{Testing}\\
\cmidrule(lr){2-5}\cmidrule(lr){6-8}
Method & GPU-h & kWh & kg CO$_2$e & Water (L) & GPU-h & kWh & kg CO$_2$e\\
\midrule
\grp{8}{Frame-level}
ForAda & 27.4 & 8.13 & 0.526 & 14.6--32.5 & 4.2 & 1.22 & 0.079 \\
Effort & 14.2 & 3.96 & 0.256 & 7.1--15.8 & 3.3 & 0.73 & 0.047 \\
GenD & 25.5 & 7.48 & 0.484 & 13.5--29.9 & 4.3 & 1.15 & 0.075 \\
DFD-HR & 86.1 & 24.80 & 1.605 & 44.6--99.2 & 9.3 & 2.75 & 0.178 \\
\midrule
\grp{8}{Clip-level}
UNITE & 17.8 & 5.71 & 0.370 & 10.3--22.8 & 7.2 & 2.22 & 0.144 \\
DeMamba & 38.6 & 9.74 & 0.630 & 17.5--39.0 & 3.3 & 0.74 & 0.048 \\
ReStraV & 13.6 & 2.57 & 0.166 & 4.6--10.3 & 3.1 & 0.60 & 0.039 \\
\midrule
\rowcolor{oursrow}
\textbf{MARS (Ours)} & 17.4 & 5.46 & 0.353 & 9.8--21.8 & 5.5 & 1.61 & 0.104 \\
\bottomrule
\end{tabular}
}
\end{table}

\FloatBarrier